\documentclass{article}
\usepackage[utf8]{inputenc}
\usepackage{palatino}
\usepackage{graphicx}
\graphicspath{ {images/} }
\usepackage[labelfont=bf]{caption}
\usepackage{subcaption}
\usepackage{setspace}
\usepackage[margin=1in]{geometry}
\usepackage[hidelinks]{hyperref}
\usepackage{xcolor}
\usepackage{overpic}
\usepackage[all]{nowidow}
\usepackage{algorithm}
\usepackage{algpseudocode}

\usepackage[disable]{todonotes}
\usepackage{array}
\newcolumntype{P}[1]{>{\centering\arraybackslash}p{#1}}
\usepackage{amsmath}
\usepackage{amsthm}
\usepackage[thinc]{esdiff}
\usepackage{amssymb}
\usepackage{float}

\DeclareMathOperator*{\argmin}{arg\,min}

\usepackage[
backend=biber,
style=authoryear,
doi=false,
url=false
]{biblatex}
\usepackage{dsfont}
\usepackage{url}

\usepackage[flushleft]{threeparttable}
\usepackage{wrapfig,lipsum}
\usepackage{booktabs, multicol, multirow,rotating}
\newtheorem{theorem}{Theorem}

\newtheorem{definition}{Definition}

\title{\vspace{-2.0cm} Local Cluster Cardinality Estimation for Adaptive Mean Shift}
\author{Étienne Pepin (\url{etienne.pepin78@gmail.com)}}

\begin{document}
\maketitle

\begin{abstract} 
This article presents an adaptive mean shift algorithm in which every parameter used at a point is derived from that point's own distance distribution. The distance distribution from a point to all others is used to estimate the cardinality of the local cluster by identifying a local minimum in the density of that distribution; the statistics of the identified subset then set the bandwidth and the kernel radius threshold applied at that point. The estimator built this way is \emph{scale invariant}, since the $\gamma$ function it rests on is unchanged when the data is multiplied by a positive constant, so no length constant has to be chosen for the scale of the data. It is also \emph{local}: $\gamma$ evaluated at rank $k$ depends only on the $k$ nearest distances, and the mean shift kernel is truncated at the estimated cluster radius, so data lying beyond that radius neither enters the estimate of the local cluster nor contributes to the weighted mean. This contrasts with kernel density estimation, which in its basic form measures density in a neighborhood of fixed size and needs a bandwidth chosen for the dataset as a whole. Our algorithm is competitive within the adaptive mean shift family: it obtains a higher Rand index than the weighted adaptive mean shift method of \cite{ren_weighted_2014} on seven of the nine datasets of that study, four of them by more than $0.03$ and three by less than $0.012$, and it performs competitively on a broader clustering benchmark, in both cases without being given the number of clusters.
\end{abstract}\hspace{10pt}

\section{Introduction}
The mean shift algorithm moves all data points towards a local average. Many methods have been implemented over the years to determine which points contribute to this local average. The most basic one simply requires a bandwidth parameter (\cite{comaniciu_mean_2002}), while adaptive mean shift methods, such as \cite{comaniciu_variable_2001}, change the bandwidth parameter locally to adapt to local data, often with a local density estimation. In this paper, we introduce a method that estimates the cardinality of local clusters and uses this information to compute cluster parameters directly, rather than relying on local density estimation. Two properties of this construction motivate it, and we return to both throughout the article. The first is \textbf{scale invariance}: the $\gamma$ function used to locate the cluster boundary is a ratio of a variance to a squared distance, so multiplying the data by a positive constant leaves it unchanged (Section \ref{sec:gammaDetails}). The estimator therefore takes no bandwidth and no distance threshold; the two constants that remain elsewhere in the algorithm, the merge threshold and the convergence tolerance of the mean shift loop, are given in Section \ref{sec:ams}. The second is \textbf{locality}: $\gamma$ evaluated at rank $k$ is a function of the $k$ nearest distances alone, so the portion of the curve that describes the local cluster cannot be perturbed by data lying further away; and the mean shift kernel is truncated at the estimated cluster radius, so those distant points also receive zero weight in the weighted mean. The parameters used at a point are therefore read from that point's neighborhood and expressed in the units of that neighborhood. We compared our method with the adaptive mean shift algorithm of \cite{ren_weighted_2014} and obtained a higher Rand index on seven of the nine datasets of that study, and our method performs competitively on a general clustering benchmark. The code used to produce those results is available  online\footnote{\url{https://github.com/pEtienn/Adaptive-mean-shift-based-on-local-cluster-cardinality-estimation}}.

The local cluster cardinality is estimated using distance distributions, which are the distribution of distances from one point to other points. Our algorithm searches for the minimum density between the leftmost mode (representing the local cluster) and the rightmost mode of the distance distributions. In our model, this point delineates the boundary of the local cluster and determines its cardinality. The estimated cardinality is then used to calculate statistics on the local cluster, such as its distance variance. Finally, this information is used during the mean shift process to adapt the bandwidth and define a distance threshold for selecting points to include in the local average.

We tested our algorithm against the results of the weighted adaptive mean shift algorithm of \cite{ren_weighted_2014} and obtained a higher Rand index on seven of the nine datasets they used, four of them by a clear margin and three by less than $0.012$. Within the adaptive mean shift family this makes our method competitive, with the caveat that the comparison covers a single member of that family: we report it against the one adaptive mean shift algorithm whose published results we could compare against directly on identical data and preprocessing. We also compared our algorithm with the benchmarks from \cite{gagolewski_framework_2022}, and the results were mixed, with our algorithm performing better on some datasets. One distinct advantage our algorithm has over the algorithms included in this benchmark is that it does not require the number of clusters as an input.
\section{Distance Distributions}
A distance distribution is the distribution of distances from one point to other points. Distance distributions can be derived mathematically for a particular distribution. For example, the isotropic Gaussian distribution in $d$ dimensions has a chi distribution with $d$ degrees of freedom (see Section \ref{sec:euclidean_distance} in the Appendix). Figure $\ref{fig:chi2D}$ illustrates how the chi Cumulative Distribution Function (CDF) is derived by integrating a Gaussian density over a sphere of a given radius in $d$ dimensions (a circle, in the two-dimensional example of the Figure).

\begin{figure}[H]
    \centering
    \includegraphics[width=0.5\linewidth]{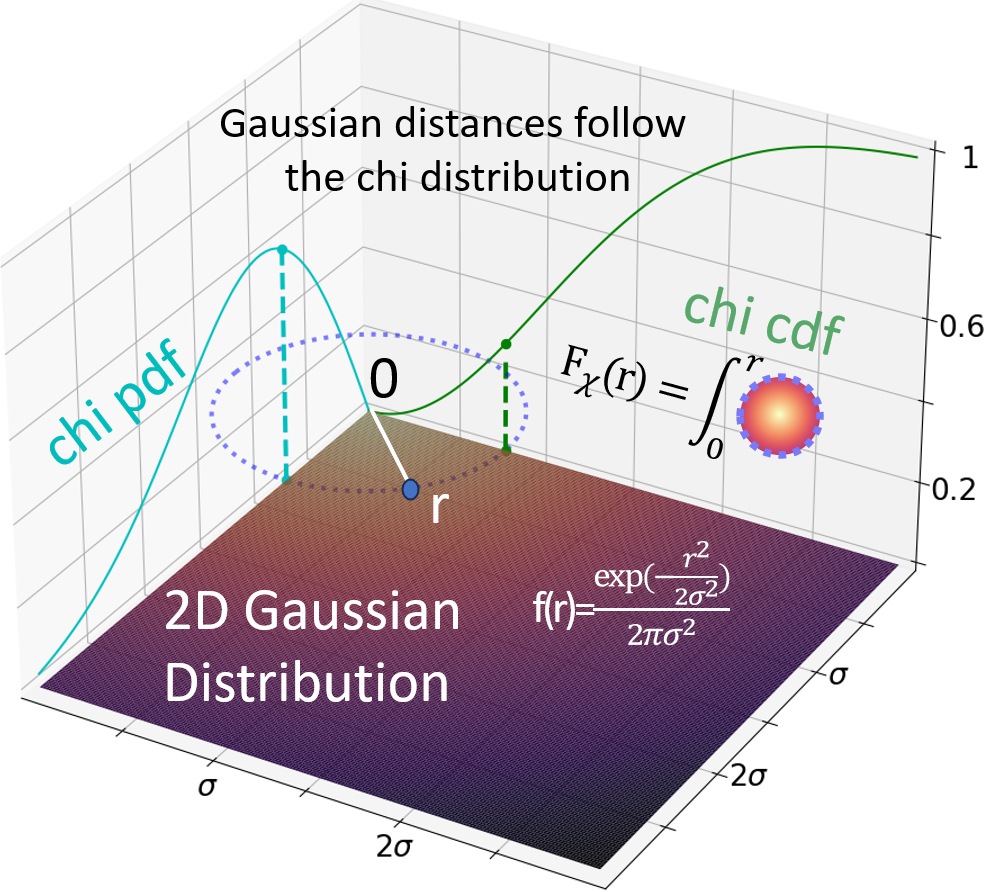}
    \caption{Distance distribution visualization}
    \label{fig:chi2D}
\end{figure}

Figure \ref{fig:chi2D} is also representative of the way many distance distributions behave. As the number of dimensions tends towards infinity, the distance distributions of independent and identically distributed (i.i.d.) distributions tend towards a normal distribution (see Section \ref{appendix:dim} in the Appendix).

Our example above considered a distance distribution between the center point of the distribution and the points it generated. A similar concept applies to distance distributions between any pair of points, as described below.

Let $Y$ be the distance distribution between the random variable vector $\mathbf{V}$ and $\mathbf{Q}$ such that $Y=\mathbf d(\mathbf{V},\mathbf{Q})$. This is the distribution of distances between two points sampled at random, which is defined for the Euclidean squared distance by the following:
\begin{definition}
    Let $\mathbf{V}$ and $\mathbf{Q}$ be two d-dimensional distributions. Then, the squared Euclidean distance distribution $Y^2$ between the 2 random points $\mathbf{V}$ and $\mathbf{Q}$ is:
\begin{equation}
    Y^2=||\mathbf{V}-\mathbf{Q}||^2=\sum_i(V_i-Q_i)^2
\end{equation}
\end{definition}
which is the definition used in \cite{angiulli_behavior_2018}. Assuming $\mathbf{V}$ and $\mathbf{Q}$ are independent and follow the same distribution, then the expectation of $Y^2$ is twice the trace of the covariance matrix of $\mathbf{V}$, $\mathrm{E}[Y^2]=2\operatorname{tr}(\mathrm{Cov}(\mathbf{V}))$, among other properties. Using this definition, we show that normal distributions also have a distance distribution between points that follow the chi distribution, as shown in Section \ref{sec:euclidean_distance} in the Appendix. 

\subsection{Distance Distributions of a Dataset}
Let $\mathcal{X}=\left\{\mathbf{X}_{1}, \mathbf{X}_{2}, \ldots, \mathbf{X}_{n}\right\}\ |\ \mathbf{X}_i=[ X_{i,1},\dots,X_{i,d}]$ denote the dataset. The ordered distance distribution from each random sample to all the others is: 

\[
 \begin{bmatrix}
    \mathbf{Y}_1\\
    \vdots\\
    \mathbf{Y}_n
\end{bmatrix}=
\begin{bmatrix} 
    Y_{1,(1)} & \dots& Y_{1,(n_1)}&\dots &Y_{1,(n-1)}\\
    \vdots &  &   &&\vdots\\
    Y_{n,(1)} &  \dots   & Y_{n,(n_n)} &\dots & Y_{n,(n-1)}
    \end{bmatrix}
,\]
where $Y_{1,(1)}\leq Y_{1,(2)}\leq \dots\leq Y_{1,(n_1)}\leq Y_{1,(n-1)}$ and $\mathbf{Y}_1=\text{sort}([\ \parallel\mathbf{X}_1-\mathbf{X}_2\parallel,\parallel\mathbf{X}_1-\mathbf{X}_3\parallel,\dots,\parallel\mathbf{X}_1-\mathbf{X}_n\parallel\ ])$. There are $n$ distance distributions $\mathbf{Y}_i$, each containing a local distance distribution of varying unknown length $n_i$ generated by a local cluster, followed by distances to other clusters. $n_i$ is unknown and $Y_{i,(j)}$ stands for the $j$th order statistic of the $i$th distance distribution from $\mathbf{X}_i$ and is always positive.

In practice, some points from other clusters may be closer than certain points within the same cluster. We will introduce some strategies to mitigate this problem. 

\section{Local Cluster Cardinality Estimator}\label{sec:clusterCardEst}
We define the cluster cardinality estimation task as follows: given a distribution of ordered distances $\mathbf{Y}$ from a point $\mathbf{x}$, estimate the cardinality $\hat n$ of the cluster to which $\mathbf{x}$ belongs.
In this section, we present our new method, which was used in our experiments, followed by potential improvements.

\subsection{Our Method: Minimum of the $\gamma$ Function}
\begin{figure}[H]
    \centering
    \includegraphics[width=0.7\linewidth]{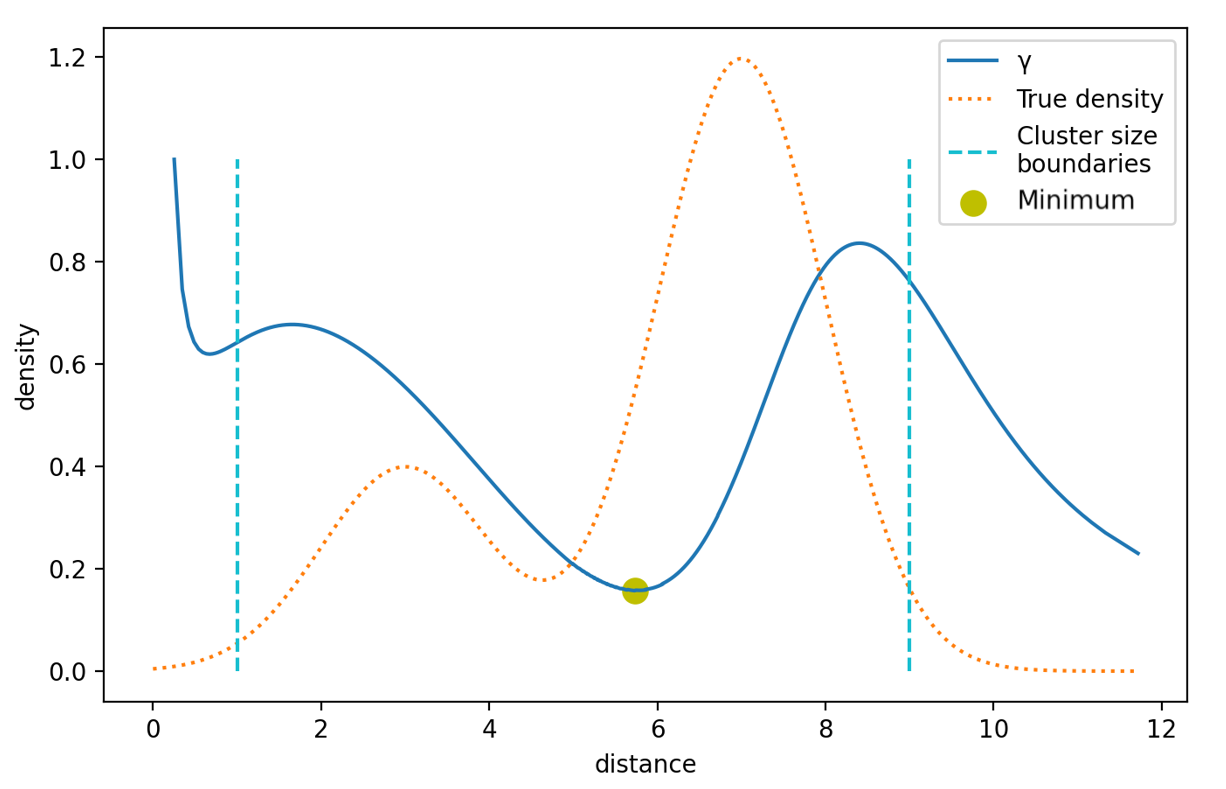}
    \caption{Cluster cardinality estimation}
    \label{fig:clusterSizeEstimation}
\end{figure}

To simplify this task, we make the hypothesis that in the context of mean shift clustering, the distance distributions from each observation can be approximated by bimodal distributions. Our algorithm aims to find the minimum density between the two modes, whose position serves as our estimate of the local cluster cardinality. Figure \ref{fig:clusterSizeEstimation} shows this process. The samples were generated from a bimodal distribution, and we apply the $\gamma$ function to estimate its true density.

In Figure \ref{fig:clusterSizeEstimation}, the minimum of $\gamma$ is found at the $i$th ordered sample of the distance distribution, which is then taken as the local cluster cardinality estimate. The definition of the $\gamma$ function is:

\begin{equation}\label{eqn:gamma}
    \gamma (\mathbf y_{(k)})=\frac{s^2}{(\bar y-y_{(k)})^2},
\end{equation}

where $\mathbf y_{(k)}=[y_{(1)},\dots,y_{(k)}]$, $\bar y$ is the sample mean and $s^2$ is the sample variance
\begin{equation*}
    \bar y= \sum_1^k y_{(i)}/k \quad \text{and}\quad s^2= \sum_1^k (y_{(i)}-\bar y)^2/k.
\end{equation*}

We calculate $\gamma$  for all distance vectors:
\[\begin{bmatrix} 
    \gamma_{1,(1)} &\dots& \gamma_{1,(n_1)} & \dots &\gamma_{1,(n-1)}\\
    \vdots   &  & &&\vdots\\
    \gamma_{n,(1)}  &\dots   & \gamma_{n,(n_n)} &\dots & \gamma_{n,(n-1)}
    \end{bmatrix}:=
\begin{bmatrix}
    \gamma(\mathbf{y}_{1,(1)}) & \dots& \gamma(\mathbf{y}_{1,(n-1)})\\
    \vdots&&\vdots\\
    \gamma(\mathbf{y}_{n,(n)})& \dots& \gamma(\mathbf{y}_{n,(n-1)})
\end{bmatrix}
\]

and estimate the cluster cardinality with
\[
\hat n_i =\argmin [\gamma_{i,(minBoundary)} \dots\gamma_{i,(n_i)} \dots \gamma_{i,(maxBoundary)}],
\]
where $minBoundary$ and $maxBoundary$ are parameters of the cluster size estimation algorithm (see Algorithm \ref{alg:clusterSizeEstimation} in the Appendix).
The estimated cluster radius for the $i$th point will be equal to $y_{\hat n_i}$, the $n_i$th distance. Since the vector $\mathbf{y}_{(\hat n_i)}=[y_{(1)},\dots,y_{(\hat n_i)}]$ contains only distances generated from the local cluster, other useful statistics can be computed on it, including sample variance, mean, percentiles and mean square distance
\begin{equation*}
    MSD=\frac{1}{d\hat n_i}\sum y_{(i)}^2,
\end{equation*}
where $d$ is the number of dimensions.

Kernel Density Estimation (KDE) methods on the data space are often used in adaptive mean shift, but cluster cardinality estimation has some advantages over it. Cluster cardinality estimation allows estimating statistics restricted to the cluster, while KDE methods measure the density of the data space locally, often using a small fraction of the cluster. For example, let both methods be used on a multivariate i.i.d. Gaussian distribution: our method can estimate the MSD, which for a Gaussian distribution recovers the factor $\sigma^2$ of its diagonal covariance $\sigma^2 I$ up to a known constant. It is equal to $\sigma^2$ for distances measured from the center of the distribution and to $2\sigma^2$ for distances between pairs of points, the case the algorithm uses, since by Theorem \ref{theo:point-to-point_euclidean} the interpoint distances follow $\chi(d,\sigma\sqrt2)$. KDEs estimate $f(\mathbf{x})$, the density of the Gaussian distribution at point $\mathbf{x}$.

\subsubsection{Details on the $\gamma$ Function and the Minimum Detection}\label{sec:gammaDetails}
 The $\gamma$ function has some properties that make it helpful in estimating the minimum location between two modes compared with 1D KDEs. Experimentally, the $\gamma$ function's minimum value, before the second mode, is near the true location of the minimum density between the two modes. In that range, it behaves like a KDE and has some advantages over it. The $\gamma$ function has high values before the first mode and is scale-invariant. It has been used in the Maximum Likelihood Estimator (MLE) of a truncated normal in \cite{cohen_simplified_1959}, and to our knowledge, it is the first time it has been used to find local minima of densities. 

The $\gamma$ function answers two problems that might be hard to solve with KDEs.
\begin{enumerate}
    \item The minimum might be found before the first mode.
    \item The bandwidth parameter of the KDE is non-trivial to choose.
\end{enumerate}

The function $\gamma$ addresses the first problem by having high values before the first mode.  For instance, when there are only two values, $\gamma(\mathbf y_{(2)})=1$, which provides an inertia-like effect before reaching the first mode.

The second problem is inherently resolved by using $\gamma$,  as it does not require a bandwidth parameter. The scale of the data does not affect the $\gamma$ function since it is scale-invariant as shown below. 
\begin{proof}
   Let $Y$ be a scale family such that $Y=\sigma B$. Then, $\gamma(\mathbf{Y}_{(k)})=\gamma(\mathbf{B}_{(k)})$ since:
\[
\gamma (\mathbf Y_{(k)})=\frac{\sigma^2 s^2(\mathbf B_{(k)})}{\sigma^2(\bar B-B_{(k)})^2}
.\]
\end{proof}

A third property, which we will refer to as locality, follows directly from Eq. \ref{eqn:gamma}: $\gamma(\mathbf y_{(k)})$ is computed from $y_{(1)},\dots,y_{(k)}$ only. Distances beyond rank $k$ do not appear in it. Consequently, if the dataset is modified beyond the local cluster, by adding, removing or displacing points that are farther from $\mathbf{x}_i$ than its own cluster, the values of $\gamma_{i,(k)}$ for every $k$ up to the local cluster's cardinality are unchanged. Such a modification can only add candidate minima at higher ranks; it cannot alter the part of the curve that describes the local cluster.

The rejection test of Algorithm \ref{alg:clusterSizeEstimation} does not inherit this property. It discards a point whose minimum sits at the edge of the search window, which happens when that point's cluster is large relative to the maximum boundary. Since the maximum boundary is a fraction of $n$, the decision to keep or to discard a point depends on the size of the whole dataset even though the values of $\gamma_{i,(k)}$ do not: adding distant points raises the boundary and can move a point from discarded to kept without changing any of its distances below rank $k$. The test asks whether a cluster is large relative to the boundary, not whether a minimum is local.

The boundary parameters are used to restrict the minimum finding to the region between the two modes. Typically, the lowest density of a bimodal distance distribution occurs near the right extremity, when distances tend towards infinity. The maximum boundary prevents the algorithm from incorrectly identifying a minimum in the right tail. Its precise value is not critical, as long as it excludes densities lower than the minimum between the two modes. 

The $\gamma$ function has not been extensively tested as a KDE. Further experiments and direct comparisons with KDEs are warranted.

\subsection{Improvements}
\subsubsection{Mode Finding}
Instead of setting boundaries via fixed parameters, they could be determined algorithmically. The cardinality estimation algorithm performs best when the two boundaries are located near the modes of the bimodal distance distribution. Those modes can be found separately using mode-finding algorithms, such as mean shift, applied directly to the distance distributions. 

For the rightmost mode (distances to points outside the cluster), the mean shift can be initiated with a single seed placed in the right tail, and the bandwidth can be set using the method of \cite{silverman_density_1998}:

\begin{equation}
\begin{split}
    h&=0.9An^{-1/5}\\
    A&=\text{min}(\text{standard deviation, interquartile range}/1.34) 
\end{split}
\end{equation}

For the left mode (distances to the same cluster), the mean shift can be initiated with one seed in the left tail. The appropriate bandwidth is often harder to determine in this case. For the right boundary case, a bandwidth computed on the whole set of distances tends to be adequate because the right mode usually contains the most points. For the left mode, particularly when it is generated by distances from a very small cluster compared with the whole dataset, computing the bandwidth on the whole dataset can result in over-smoothing and possibly missing the mode. Adaptive mean shift methods could be considered to avoid this issue. The leftmost mode appears less critical than the rightmost mode to locate precisely, since the $\gamma$ function tends to yield high values before the first mode. Therefore, improvements should focus primarily on accurately locating the rightmost mode.

\subsubsection{Kernel Density Estimation}
We should also consider replacing the $\gamma$ function with a kernel density function to locate the point of minimum density. The difficulty is the same as for finding the left mode. For very small clusters, it is easy to oversmooth the density and then possibly miss a minimum in density.

Finding an appropriate bandwidth for the small distances would also enable more sophisticated methods than locating the minimum between two boundaries. For instance, we could discard the bimodal assumption and instead identify all modes of the distance distribution. In this case, the estimated cardinality would correspond to the minimum density between the first two modes. This approach is more complex and more dependent on finding a suitable bandwidth, but it may perform better in some cases where the current algorithm fails.

\subsubsection{Task Definition and Performance Evaluation}
To our knowledge, local cluster cardinality estimation is a novel task. Evaluating performance on this task is challenging, as it is intended to be applied to a broad range of multimodal distributions. Consequently, our performance analyses are often exploratory in nature. Clearly defining the task and establishing robust evaluation criteria are important priorities moving forward.

\section{Adaptive Mean Shift}\label{sec:ams}
The mean shift procedure identifies modes by iteratively computing the weighted mean (Eq. \ref{eq:weightedMean}) of points within a local window, shifting the window to this mean, and repeating the process until convergence. The weighted mean equation is shown below:
\begin{equation} \label{eq:weightedMean}
    m(\mathbf{x},\mathcal{X}) = \frac{ \sum_{\mathbf{x}_i \in \mathcal{X}} K(||\mathbf{x}_i - \mathbf{x}||) \mathbf{x}_i } {\sum_{\mathbf{x}_i \in \mathcal{X}} K(||\mathbf{x}_i - \mathbf{x}||)}.
\end{equation}
The adaptive part in our procedure is defined by the Gaussian Kernel $K$, which uses information from local distance distributions to adapt the window size and the bandwidth locally.

\subsection{Gaussian Kernel}
We tested two variations of the Gaussian kernel that utilize information from parameter estimation on distance distributions: a Gaussian kernel and a Gaussian kernel adapted for high dimensions. The Gaussian kernel is presented below:
\large\begin{equation}\label{eq:gaussK}
K(x;h,\omega) =
\begin{cases}  
 \exp{\left(-\frac{x^2}{2 h^2}\right)}\ & \text{if}\ x \le \omega\\
0 & \text{if}\ x > \omega\\
\end{cases} 
\end{equation},
\normalsize
where $x$ is the distance, $\omega$ is the distance threshold and $h$ is the bandwidth. The distance threshold $\omega$ is equal to $y_{(\hat n_i)}$, the $\hat n_i$th ordered distance from point $i$. The cluster cardinality limit ($\hat n_i$) aims to restrict contributions to the weighted average to points within the same cluster. This prevents a large number of distant points with small contributions from influencing the local mean more than a small number of points from the local cluster, which can occur in datasets containing small clusters. It is also how the locality of Section \ref{sec:gammaDetails} carries into the mean shift step: the kernel is exactly zero beyond $\omega$, so points outside the estimated local cluster contribute nothing to the weighted mean.

Currently, we use the standard deviation of the local distance distribution as the bandwidth. Another possibility is to use a method from \cite{silverman_density_1998}, based on minimizing the mean integrated square error (p.85-87). The one-dimensional version of this method is widely used, for example, being part of the default bandwidth estimator for kernel density estimation in scikit-learn (\cite{pedregosa_scikit-learn_2011}). However, it does not seem that the high-dimensional version of this method is used as much, and it has performed worse in our tests than the standard deviation of the distance distribution.

Next, we show our Gaussian kernel adapted for high dimensions:
\large
\begin{equation}\label{eq:gaussHD}
K(x;h,\omega,a,s,\bar x) =
\begin{cases}  
 \exp{\left(-\frac{(\max [x-(\bar x-as),0])^2}{2 h^2}\right)}\ & \text{if}\ x \le \omega\\
0 & \text{if}\ x > \omega\\
\end{cases} 
\end{equation}
\normalsize
where $\bar x=\frac{1}{\hat n_i}\sum_{i=1}^{\hat n_i}x_i$, $s^2=\frac{1}{\hat n_i -1}\sum_{i=1}^{\hat n_i}(x_i-\bar x)^2$, $a\in\mathbb{R}^+$ and $\{x_1,\dots,x_{\hat n_i}\}$ is the local distance sample. In our experimentation, $a$ is set to 4 to bring the smallest distances to a value slightly above or equal to zero.

Equation \ref{eq:gaussHD} is a variation on Eq. \ref{eq:gaussK} that offsets distances closer to zero. The aim is to avoid two issues that arise in high-dimensional spaces, where dimensions are usually a lot larger (see Section \ref{appendix:dim} in Appendix). 
The first problem is that on the first iteration of the mean shift algorithm, the distance between a point and itself is zero, which assigns it an excessively large weight relative to other points. The algorithm could get stuck at this point or simply take a lot of time to move the initial weighted mean positions away from the original points. Another advantage of this kernel is that it helps prevent computing very small values when distances are very large. Similar to Eq. \ref{eq:gaussK}, the $\omega$ threshold ensures that kernel contributions come predominantly from points within the same cluster—likely the most significant enhancement we provide for clustering in high-dimensional settings.

\newpage
\subsection{Cluster Cardinality-Based Adaptive Mean Shift}
Below is our adaptive mean shift algorithm. 

\begin{algorithm} [H]
\small
\caption{Cluster Cardinality-Based Adaptive Mean Shift}\label{alg:ams}
 \hspace*{\algorithmicindent} \textbf{Input} $\mathcal{X},(minBoundary\gets 5, maxBoundary\gets n/2),maxIter\gets 250$\\
 \hspace*{\algorithmicindent} \textbf{Output} $\mathcal{M},\mathbf{label}$
\begin{algorithmic}
\State \textbf{Step1:} Cluster Cardinality Estimation
\State $\mathcal{P}_1,(\mathbf{\hat n},\mathbf{h})\gets \text{clusterCardinalityEstimation}(\mathcal{X},(minBoundary, maxBoundary))$ \Comment{See algorithm \ref{alg:clusterSizeEstimation} in the Appendix}
\vspace{5pt}
\State \textbf{Step2:} Adaptive Mean Shift
\For {$j: 1 \rightarrow maxIter$} 
\For{$\mathbf{p}_{j,i} \in \mathcal{P}_j$}
\State $[0,y_{(1)},\dots,y_{(n_1)}]\gets\text{sort}([\ \parallel\mathbf{x}_i-\mathbf{x}_1\parallel,\dots,\parallel\mathbf{x}_i-\mathbf{x}_{n_1}\parallel\ ])$
\State $\hat n_k \gets$ getLocalNParameter($\mathbf{p}_{j,i}$)
\State $n_{k*}\gets \hat n_k\text{min}[ (minBoundary+0.01j*(\hat n_k-minBoundary))/\hat n_k,1]$ \Comment{area increase, see section \ref{sec:gradualIncrease}}
\State $h=\text{Std}(y_{(1)},\dots,y_{(n_{k*})})$
\State $\omega=y_{(n_{k*})}$
\vspace{4pt}
\large
\State $\mathbf{p}_{j+1,i} \gets \frac{ \sum_{\mathbf{x}_i \in \mathcal{X}} K(||\mathbf{x}_i - \mathbf{p}_{j,i}||;h,\omega) \mathbf{x}_i } {\sum_{\mathbf{x}_i \in \mathcal{X}} K(||\mathbf{x}_i - \mathbf{p}_{j,i}||;h,\omega)}$
\small
\vspace{4pt}
\EndFor
\State $\mathcal{P}_{j+1}\gets[\mathbf{p}_{j+1,1},\dots]$
\State break if $\mathcal{P}_j\approx \mathcal{P}_{j+1}$ and $j>100$ \Comment{Summed movement from $\mathcal{P}_j$ to $\mathcal{P}_{j+1}$ below $\epsilon$}
\State $\mathcal{P}_{j+1} \gets\text{merge}(\mathcal{P}_{j+1},\delta)$ \Comment{Shifted points closer than $\delta$ are combined}
\EndFor
\State $\mathcal{M},\mathbf{label} \gets$ clustering$(\mathcal{P}_j)$ \Comment{See algorithm \ref{alg:clustering} in the Appendix}
\vspace{4pt}
\State \textbf{Step3:} Classify Points With Bad Estimates
\State $\mathcal{F}\gets\mathcal{X}-\mathcal{P}_1 $ \Comment{All points with bad estimate. $-$ is the set difference operation.}
\State $\mathbf{label_{\mathcal{F}}}\gets\emptyset$
\State $\mathbf{s}^2 \gets$ getLocalKParameters($\mathcal{M}$) \Comment{Distance variance of each cluster}
\For{$\mathbf{f}_i\in\mathcal{F}$}
    \State $\mathbf{dist}\gets ||\mathbf{f}_i-\mathcal{M}||$ \Comment{Distances between $\mathbf{f}_i$ and all modes}
    \State $\mathbf{label_{\mathcal{F}}}\gets\mathbf{label_{\mathcal{F}}}\cup \text{argmin} \exp{(\mathbf{dist}^2/2\mathbf{s}^2)}$
\EndFor
\State $\mathbf{label}\gets\mathbf{label}\cup\mathbf{label}_{\mathcal{F}}$
\end{algorithmic}
\end{algorithm}
where $\mathcal{X}\in \mathbb{R}^{n\times d}$ is the original dataset, $\mathcal{P}_1\in\mathbb{R}^{n_1\times d}\ |\ n_1\leq n$ is the dataset without points with bad cardinality estimate, $\mathbf{\hat n}=[\hat n_1,\dots,\hat n_n]\ |\ n_i\in \mathbb{N}$ is the cardinality estimate for each point, $\mathbf{h}=[h_1,\dots,h_n]\ |\ h_i\in\mathbb{R}$ is the bandwidth for each point, $\mathcal{P}_{j}\in\mathbb{R}^{m_j\times d} \ |\ j>1,\ m_j\leq n_1$ is the set of distinct shifted points at iteration $j$, each carrying the labels of the points merged into it, $K$ is the Gaussian kernel, $\mathcal{M}\in \mathbb{R}^{c\times d}$ is the set of all modes and $\mathbf{label}=[label_1,\dots,label_n]\ |\ label_i\in \{1,\dots,c\}$ is the label of each point corresponding to a mode.

The algorithm takes as input a dataset $\mathcal{X}$ along with minimum and maximum boundary parameters for cluster cardinality estimation, with default values of 5 and $n/2$, respectively. If clusters bigger than $n/2$ are expected, the maximum boundary should be increased. That recommendation uses knowledge about the data; the results of Section \ref{sec:benchmark} are reported at the default, without it. The default minimum is discussed in Section \ref{sec:clusterCard}.

The mean shift loop carries two constants that the estimator does not. The merge threshold $\delta$ is the first percentile of the nearest-neighbor distances of the dataset, so it scales with the data but is a single value applied to every cluster. The convergence tolerance $\epsilon$ is compared against the summed movement of the shifted points between two iterations and is an absolute quantity, $10^{-5}$ in our implementation. Neither is read from the local distance distribution, and both should eventually be derived from the data, as proposed for the maximum boundary in Section \ref{sec:maxBoundaryEffect}.

The algorithm works in 3 steps:
\begin{enumerate}
    \item Cluster Cardinality Estimation
    \item Adaptive Mean Shift
    \item Classify Points With Bad Estimates
\end{enumerate}
In the \textbf{Cluster Cardinality Estimation} step (Section \ref{sec:clusterCardEst} and Algorithm \ref{alg:clusterSizeEstimation} in the Appendix), a cluster cardinality $\hat n_i$ is estimated for each point $\mathbf{x}_i$. This is an estimate of the number of points present in the cluster to which $\mathbf{x}_i$ belongs. Points with a cluster cardinality estimate that are affected by the maximum boundary limit are labelled "bad" and are removed from the set of points used during the mean shift. Finally, statistics are computed on the local distance distribution between $\mathbf{x}_i$ and the closest $\hat n_i$ points, such as the sample mean, variance, and MSD. The exact statistic needed depends on the kernel used during mean shift, in the above example, the Gaussian kernel (Eq. \ref{eq:gaussK}) is used. The cluster cardinality estimation step outputs the set of points with a good cardinality estimate $\mathcal{P}_1$ and the needed parameters for the Gaussian kernel (in this case, $\mathbf{\hat n}$ and $\mathbf{h}$).

The \textbf{Adaptive Mean Shift} step moves iteratively all points towards their mode. It starts with each point $\mathbf{p}_{1,i}\in\mathcal{P}_1$, computes the weighted average $\mathbf{p}_{2,i}$ in the sphere including $n_{k*}$ points around it and repeats iteratively until it reaches $\mathbf{p}_{1,i}$'s mode or reach the maximum number of iterations. 
The function "\textbf{getLocalNParameter}" takes $\mathbf{p}_{j,i}$ and returns the local space parameter $n_k$. It finds the 5 nearest points in $\mathcal{P}_1$ to $\mathbf{p}_{j,i}$ and returns the median values of the parameter $\hat{n}$ from these five points, effectively regularizing the parameters used by the adaptive mean shift. $n_{k*}$ increases progressively over 100 steps, from $minBoundary$ to $\hat{n}_k$ (see Section \ref{sec:gradualIncrease}). $h$ and $\omega$ are computed on the distance distribution $[y_{(1)},\dots,y_{(n_{k*})}]$, delineated by $n_{k*}$.

Weighted averages from Eq. \ref{eq:weightedMean} are computed at the same time to use Python's matrix operations, which is why we check if all points have stopped moving at the same time. The \textbf{clustering} function returns the modes $\mathcal{M}$ and the label of each point corresponding to the modes (see Algorithm \ref{alg:clustering} in the Appendix). Section \ref{sec:gradualIncrease} contains more details on the gradual mean shift area increase. In the code implementation, clustering is done on each iteration to reduce the size of $\mathcal{P}_{j+1}$.

Finally, the step \textbf{Classify Points With Bad Estimates} classifies all points that had a bad cluster size estimation by assigning them to the closest cluster, weighing distances by the clusters' bandwidth.

\section{Experimentations}
Experiments were conducted on a toy dataset containing four clusters designed to test the ability of the algorithm to adapt to scale and to the number of samples. 
\begin{figure}[H]
    \centering
    \includegraphics[width=0.5\linewidth]{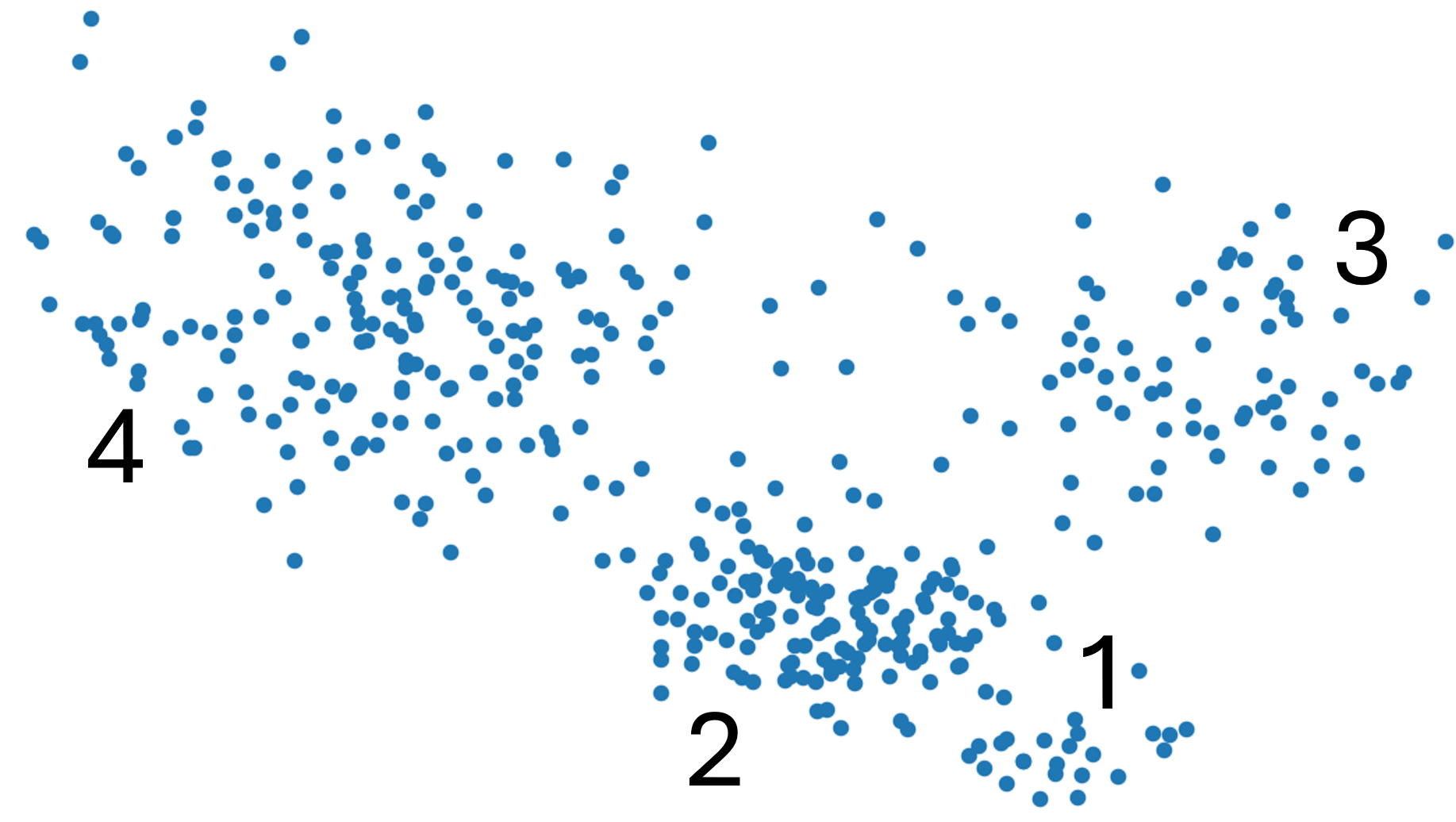}
    \caption{Toy dataset clusters}
    \label{fig:toySet}
\end{figure}
The toy dataset is generated by four isotropic Gaussian distributions with the following properties:
\begin{table}[H]
    \centering
    \begin{tabular}{c|c|c}
         cluster \# &Cardinality&$\sigma$  \\
         \hline
         1&25&0.7\\
         2&100&1\\
         3&75&1.5\\
         4&200&2
    \end{tabular}
    \caption{Toy dataset}
    \label{tab:toySet}
\end{table}
where $\sigma^2I$ is the covariance matrix for each distribution. Our algorithm used the following parameters:

\begin{table}[H]
    \centering
    \begin{tabular}{c|c}
         Parameter & Value \\
         \hline
         Minimum boundary & 5 \\
         Maximum boundary & $n/2$\\
         Maximum number of iterations &200
    \end{tabular}
    \caption{Experimentation parameters}
    \label{tab:my_label}
\end{table}
where $n$ is the total number of samples.

\subsection{Cluster Cardinality Estimation} \label{sec:clusterCard}
\begin{figure}[H]
    \centering
    \includegraphics[width=0.9\linewidth]{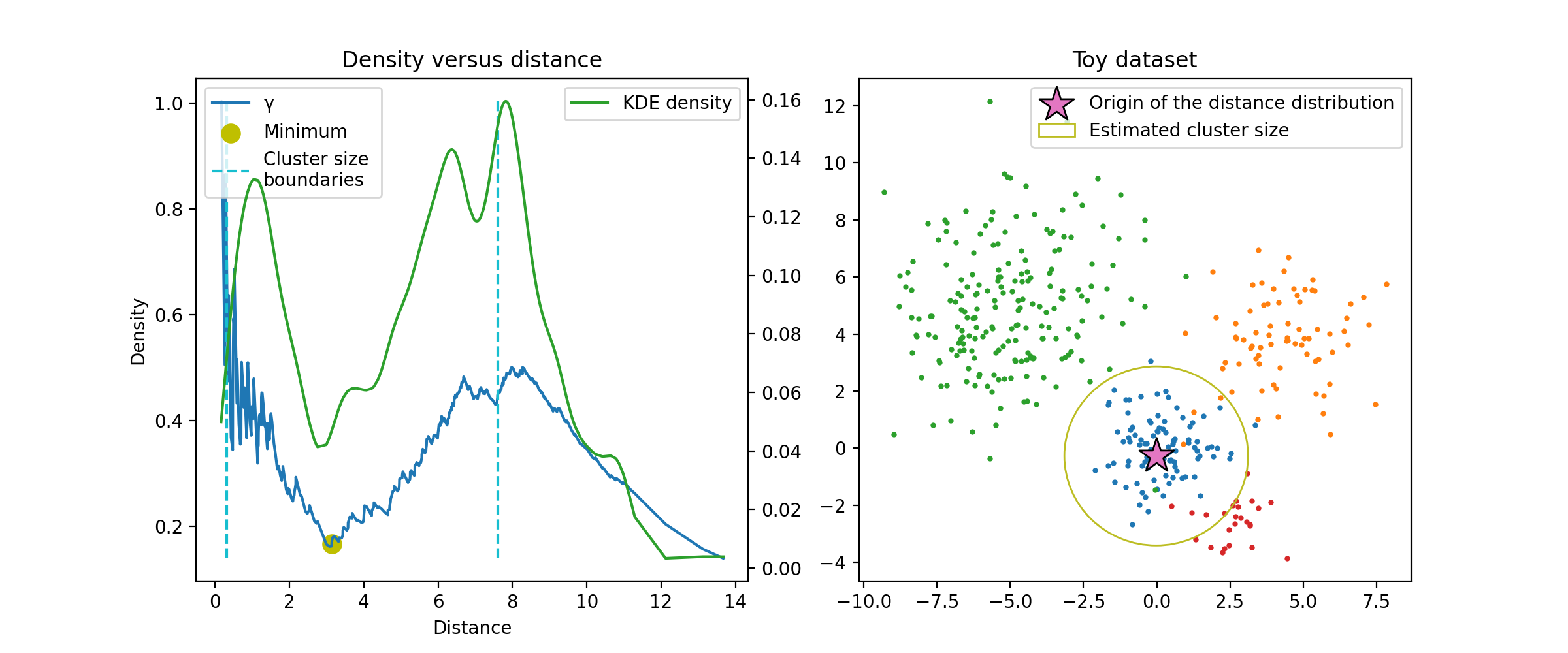}
    \caption{Cluster cardinality estimation}
    \label{fig:good Cluster size estimation}
\end{figure}

\begin{figure}[H]
    \centering
    \includegraphics[width=0.9\linewidth]{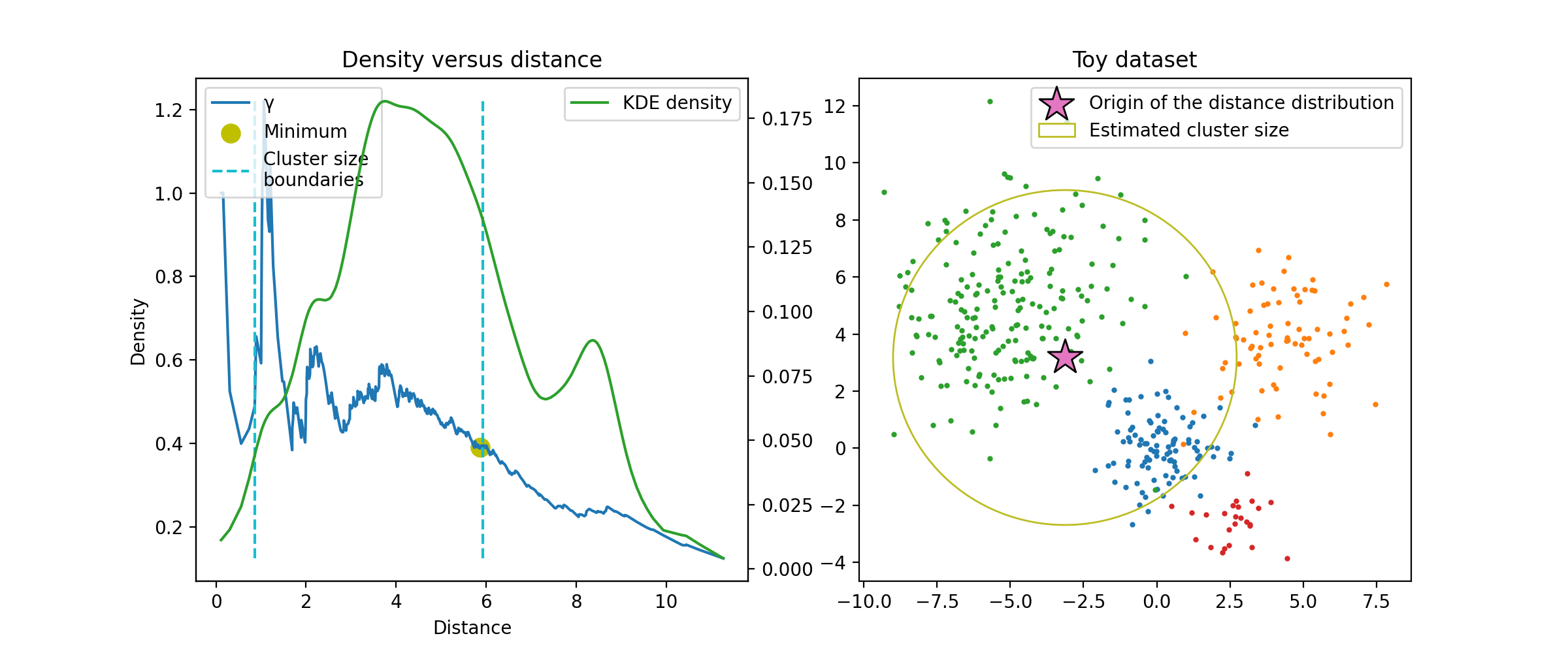}
    \caption{Bad cluster size estimation}
    \label{fig:bad cluster size estimation}
\end{figure}

Cluster cardinality is estimated for each point by finding the minimum density of the distance distribution (see Algorithm \ref{alg:clusterSizeEstimation}). Figure \ref{fig:good Cluster size estimation} shows an example with the distance distribution (left graph) from the pink star (right graph) to all other points. The estimated cluster radius ($\hat n$th distance) is indicated by the minimum marker (left graph), and we can see that it accurately delimits points in the blue cluster (on the right).  
Points near the center of a cluster generally receive more accurate cluster cardinality estimates than those located between two clusters. In the latter case, the estimate can be determined by the upper boundary limit, as shown in Figure \ref{fig:bad cluster size estimation}. We classify those estimates as "bad" and remove those points from the mean shift process.

Both $\gamma$ and KDE functions are shown on Figures \ref{fig:good Cluster size estimation} and \ref{fig:bad cluster size estimation} to show how they compare. The $\gamma$ function follows closely the behavior of the KDE function near the minimum between the two modes in Figure \ref{fig:good Cluster size estimation}, but without having low values before the first mode. This makes the exact choice of the lower boundary less critical; its primary role is to prevent the high variance typically associated with computing statistics on very small samples.

\begin{figure}[H]
    \centering
    \includegraphics[width=0.6\linewidth]{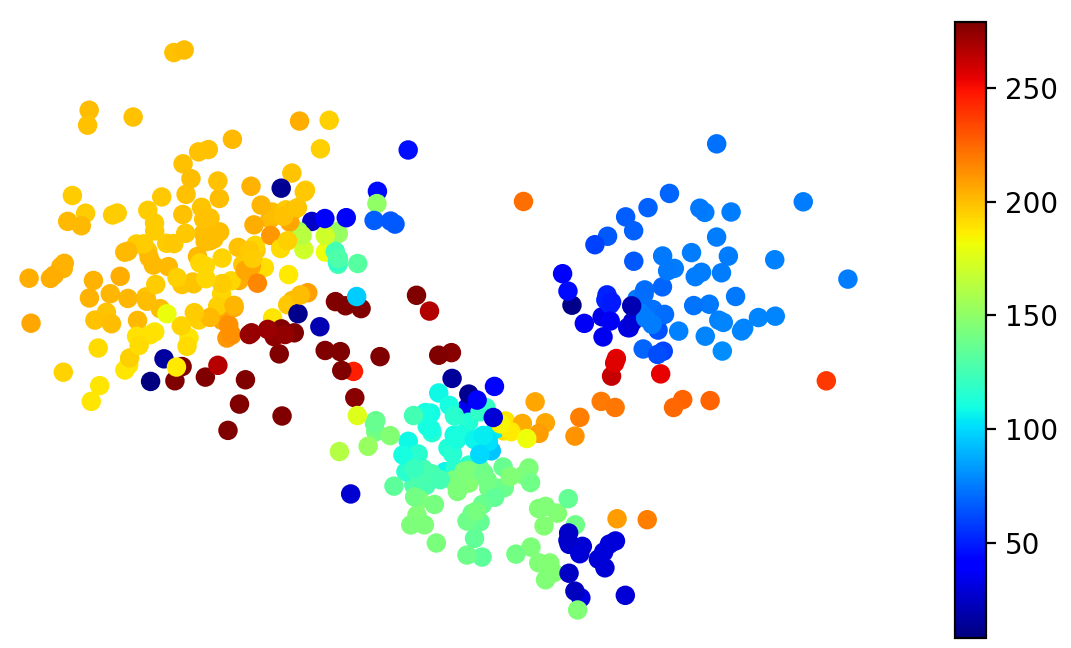}
    \caption{Cluster cardinality estimation on toy dataset}
    \label{fig:goodNestimate}
\end{figure}

Figure \ref{fig:goodNestimate} shows the estimated cluster cardinality for all points.
Cluster cardinality estimation tends to be more accurate near the center of a cluster and less reliable between clusters, due to the inclusion of nearest neighbors from adjacent clusters in the distance distribution. When these nearest neighbors belong to other clusters, it leads to an overestimation of cluster cardinality.

\subsubsection{Gradual Mean Shift Area Increase} \label{sec:gradualIncrease}
To counteract the positive bias in cardinality estimates for points far from cluster modes, we gradually increase the mean shift area during the mean shift procedure. The area considered by the kernel is gradually increased from a small value until it includes a number of points equal to the local cluster cardinality estimate. 

Points located farther from the cluster center tend to have positively biased cardinality estimates, causing their mean shift kernel to include points from neighboring clusters. This may shift the weighted mean away from its true cluster, particularly in cases where a large cluster is adjacent to a smaller one, such as clusters 2 and 1 in the toy dataset.

Increasing progressively the mean shift area might allow weighted means to gradually shift towards the right cluster mode, where the cluster cardinality estimate tends to be more accurate. Once the mean shift area reaches its full extent, the weighted means remain near the correct mode. Preliminary results with this approach were promising, although further evaluation is required.

\subsection{Clustering}
\begin{figure}[H]
    \centering
    \includegraphics[width=0.6\linewidth]{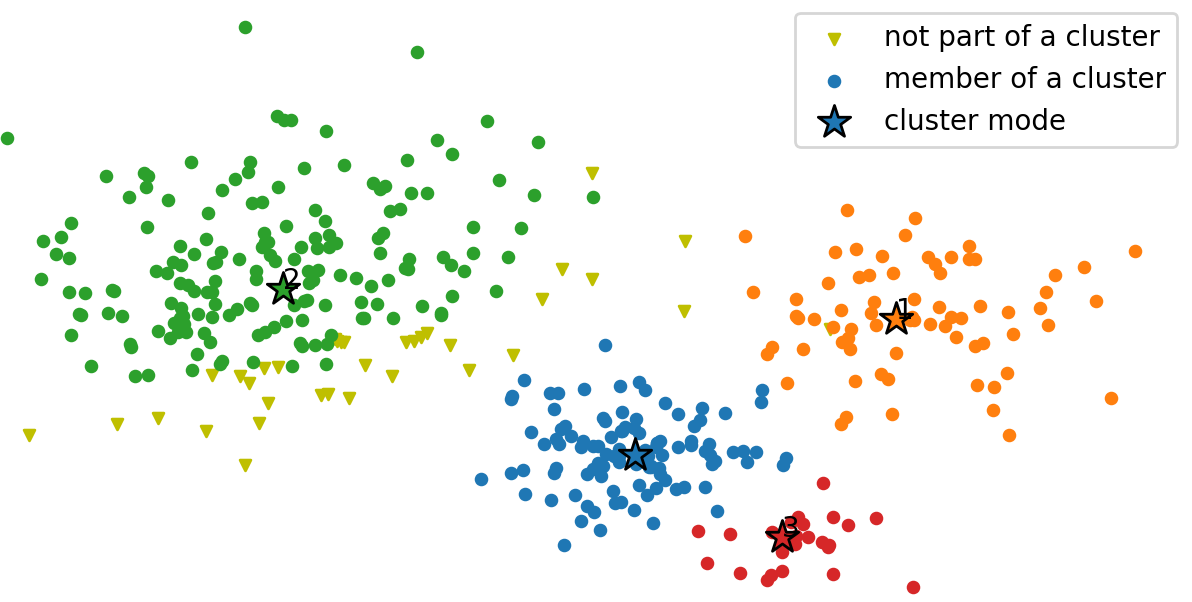}
    \caption{Clusters found on toy dataset after mean shift}
    \label{fig:clusterPreClassification}
\end{figure}

In Figure \ref{fig:clusterPreClassification}, we show clustering of the toy dataset, excluding points that have a bad cluster cardinality estimate (before Step 3 in Algorithm \ref{alg:ams}). All points marked with a triangle have a bad estimate and were not used during the mean shift. Notice how they are situated mainly between clusters.

\begin{figure}[H]
    \centering
    \includegraphics[width=0.6\linewidth]{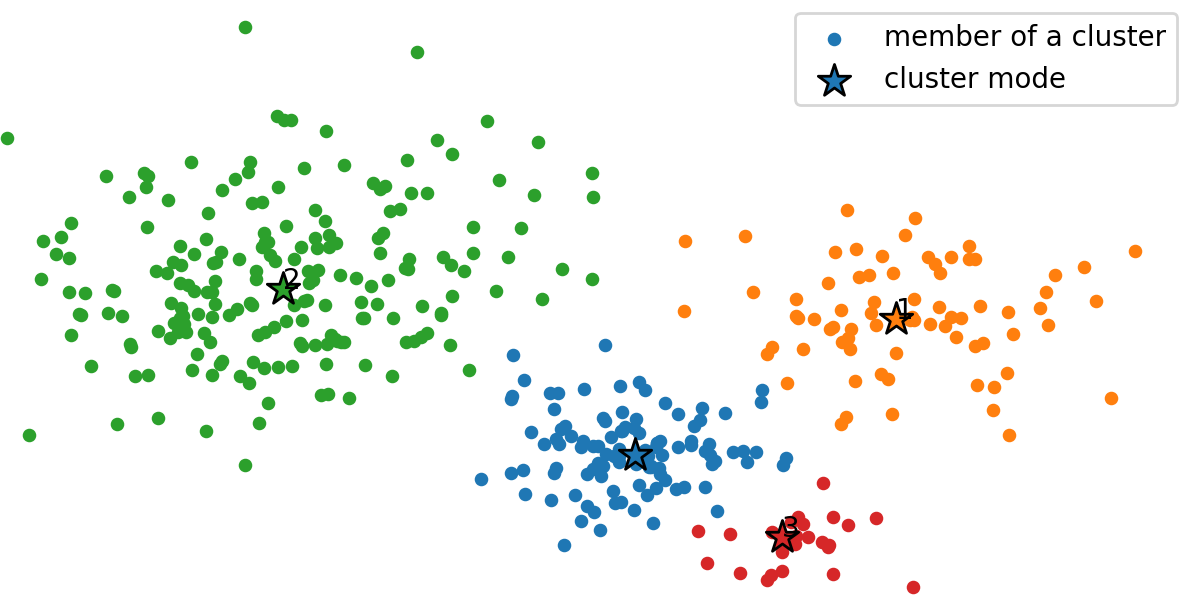}
    \caption{Clusters found on toy dataset after classification}
    \label{fig:clusterPostClassification}
\end{figure}

The classification step assigns them to the nearest cluster, accounting for the scale of the cluster, and results in Figure \ref{fig:clusterPostClassification}.

\subsection{Comparison to the Expectation–Maximization Algorithm on the Toy Dataset}
\begin{figure}[H]
    \centering
    \includegraphics[width=0.6\linewidth]{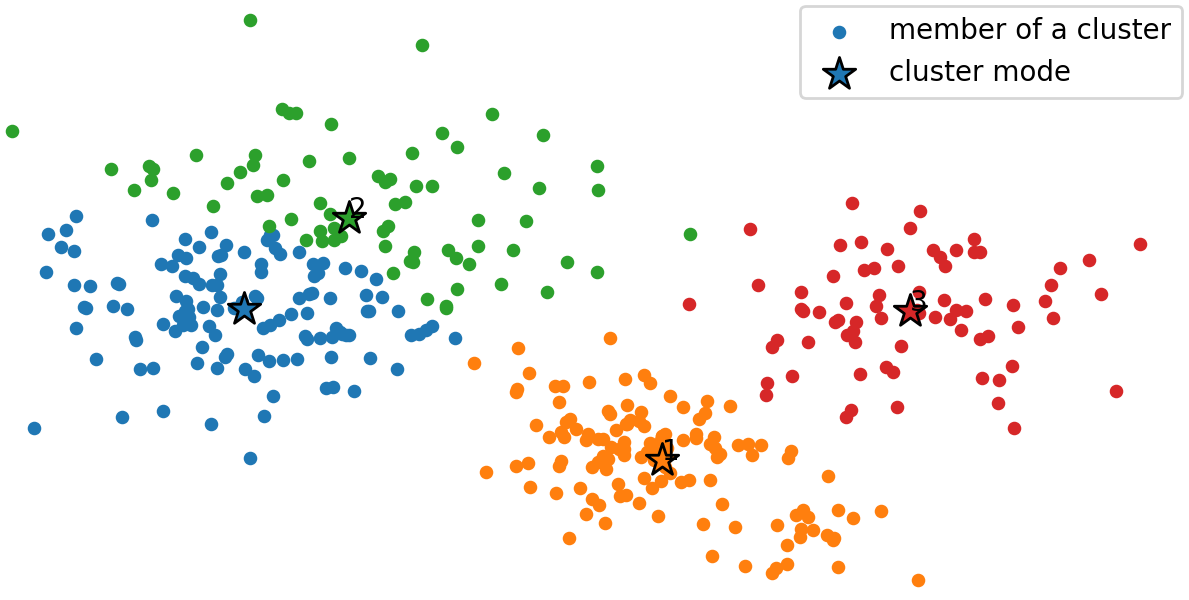}
    \caption{Clustering by the EM algorithm}
    \label{fig:em}
\end{figure}

The main advantage of our algorithm lies in its ability to handle varying scales and numbers of samples. Many clustering algorithms struggle with sets such as our toy dataset. For example, when applying the EM algorithm (\cite{dempster_maximum_1977}) on our toy dataset, it has a very high tendency to miss the small cluster, as shown in Figure \ref{fig:em}.

\section{Comparison to Other Clustering Algorithms} 
In this section, we will compare our results against a similar clustering algorithm and a benchmark library.

\subsection{Algorithm: A Weighted Adaptive Mean Shift Clustering Algorithm }
 The article \textit{Weighted Adaptive Mean Shift Clustering Algorithm} (WAMS) by \cite{ren_weighted_2014} presents an adaptive mean shift algorithm in which they "develop a technique to estimate the relevant subspace for each point, and then embed such information within the mean shift search process". We evaluated our algorithm on all datasets used in their experiments, except for the COIL dataset, which we were unable to locate in the form described in their paper.

\begin{table}[H]
\centering
\begin{tabular}{|l|ccc|}
\hline Dataset & \#points & d & \#classes \\
\hline
Iris & 150 & 4 & 3 \\
Yeast & 1136 & 8 & 3 \\
Steel & 1941 & 27 & 7 \\
USPS & 2007 & 256 & 10 \\
CTG & 2126 & 21 & 10 \\
Letter & 2263 & 16 & 3 \\
Image & 2310 & 19 & 7 \\
Pen & 3165 & 16 & 3 \\
Wave & 5000 & 21 & 3 \\
\hline
\end{tabular}
    \caption{Datasets used in the experiments of \cite{ren_weighted_2014}}
    \label{tab:dataUsed}
\end{table}
The same preprocessing as in \cite{ren_weighted_2014} was done on each of the datasets:
\begin{quote}
    USPS is a handwritten digit database and the 2007 test images were chosen for our experiments. The other eight datasets are from the UCI repository. Iris, Steel (Steel Plates Faults), CTG (Cardiotocography), Image (Image Segmentation), and Wave (Waveform version 1) are all in the original form. The three largest classes of Yeast were selected and letters 'I', 'J' and 'L' were chosen from the Letter database. Pen contains 10,992 samples from 10 classes (digits 0-9) and we selected 3 classes (digits $3,8,9)$. For each dataset, features were normalized to have zero mean value and unit variance.
\end{quote}

We used the Rand Index to evaluate performance, as was done for their experiments. The Rand Index is a measure of similarity between 2 data partitions, going from 0 to 1, which means that the two clusterings are the same. Table \ref{tab:tableRen} contains our results and results presented by \cite{ren_weighted_2014}.

\begin{table}[H]
    \centering
\begin{tabular}{|c|c|c||r|rrr|}
\hline \multirow{2}{*}{ Dataset } & \multicolumn{2}{|c|}{Cluster Cardinality-Based Adaptive Mean Shift} & \multirow{2}{*}{ WAMS }  & \multirow{2}{*}{$k$-means } & \multirow{2}{*}{ EM } & \multirow{2}{*}{ Sing-l } \\
 \cline{2-3}&Gaussian Kernel& High Dimension Kernel &&  & &\\
\hline \hline Iris &$\mathbf{0.9575}$&	0.9495& 0.8060 &  0.8051 & 0.8193 & 0.7771 \\
Yeast &$\mathbf{0.6210}$&0.6041&0.6098 & 0.5624 & 0.3713 & 0.3602 \\
Steel &0.7792	&$\mathbf{0.7794}$&$0 . 7 4 3 5$ & 0.7394 & 0.7168 & 0.2446 \\
USPS &0.7292&0.7459& $\mathbf{0 . 9 0 0 6}$ & 0.8720 & 0.8225 & 0.1274 \\
CTG &$\mathbf{0.8389}$&0.8387&$0 . 8 0 4 6$ &  0.7996 & 0.7487 & 0.2007 \\
Letter &\textbf{0.6952}&0.6925& $0 . 6 9 0 8$ & 0.6029 & 0.6074 & 0.3336 \\
Image &0.8417&0.8409&$\mathbf{0 . 8 8 2 0}$ & 0.8399 & 0.8028 & 0.1531 \\
Pen &\textbf{0.8256}&0.8125&$0 . 7 1 9 8$ &  0.7068 & 0.7143 & 0.6852 \\
Wave &\textbf{0.6761}&0.6681&$0 . 6 6 9 5$ &  0.6674 & 0.6675 & 0.6467 \\
\hline
\end{tabular}
    \caption{Comparison against clustering algorithms on
real data (Rand Index). The WAMS, $k$-means, EM and Sing-l (single-linkage hierarchical clustering) columns are the results published by \cite{ren_weighted_2014}. Bold marks the best value of each row. Each of our two columns is a single run of the algorithm, so differences in the third and fourth decimal place should not be read as differences in performance.}
    \label{tab:tableRen}
\end{table}
Our algorithm obtains a higher Rand index than WAMS on seven of the nine datasets, and a higher Rand index than every published baseline in the table on the same seven. The two exceptions are \textit{USPS} and \textit{Image}; we cannot investigate those two failures further, as we do not have access to their implementation. The margins are uneven: four of the seven lie between $0.034$ and $0.152$, on \textit{Steel}, \textit{CTG}, \textit{Pen} and \textit{Iris}, while \textit{Yeast}, \textit{Wave} and \textit{Letter} are won by $0.011$, $0.007$ and $0.004$. Each value in the table is a single run, so those last three are better read as ties than as wins.

It is worth noting that selection bias may have influenced the above results, as the datasets could have been chosen to favor the WAMS algorithm over other methods.

Our high-dimensional kernel often performs slightly worse than the Gaussian kernel, except on the USPS dataset, which has the highest number of dimensions by far. Further testing on additional high-dimensional datasets is required to confirm whether the high-dimensional kernel is indeed more effective than the Gaussian kernel in such contexts.

\subsection{Benchmark: A Framework for Benchmarking Clustering Algorithms}\label{sec:benchmark}
We also evaluated our algorithm using datasets and benchmark results published by \cite{gagolewski_framework_2022}. As before, the data is sourced from the UCI repository, although six of the eight datasets differ from those in the previous comparison. The two that recur are \textit{Statlog}, which is the Statlog (Image Segmentation) data appearing as \textit{Image} in Table \ref{tab:tableRen}, and \textit{Yeast}, of which \cite{ren_weighted_2014} used only the three largest classes.
\begin{table}[!ht]
    \centering
    \begin{tabular}{|l|ccccc|}
    \hline
        Dataset & \#points & d & \#classes & Gini index & largest class$/n$ \\ \hline
        Ecoli & 336 & 7 & 8 & 0.65 & 0.426 \\
        Glass & 214 & 9 & 6 & 0.48 & 0.355 \\
        Ionosphere & 351 & 34 & 2 & 0.28 & \textbf{0.641} \\
        Sonar & 208 & 60 & 2 & 0.07 & \textbf{0.534} \\
        Statlog & 2310 & 19 & 7 & 0.00 & 0.143 \\
        WDBC & 569 & 30 & 2 & 0.25 & \textbf{0.627} \\
        Wine & 178 & 13 & 3 & 0.13 & 0.399 \\
        Yeast & 1484 & 8 & 10 & 0.63 & 0.312 \\ \hline
    \end{tabular}
    \caption{Datasets in the UCI set by \cite{gagolewski_framework_2022}. The last column is the size of the largest reference class as a fraction of $n$; bold marks the three datasets where it exceeds the default maximum boundary of $0.5n$, which is discussed in Section \ref{sec:maxBoundaryEffect}.}
    \label{tab:benchData}
\end{table}
They use a slightly different preprocessing to \cite{ren_weighted_2014}: 
\begin{quote}
Remove all columns of zero variance (constant). Centre the data around the centroid (so that each column's mean is 0). Scale all columns proportionally (so that the total variance is 1; note that this is not the same as standardisation: standard deviations in each column might still be different). Add a tiny amount of noise to minimise the risk of having duplicate points.
\end{quote}

Table \ref{tab:benchUCIresult} presents the performance of our algorithm using the Gaussian kernel, alongside results from other clustering algorithms included in the UCI benchmark dataset.
\begin{table}[H]
    \centering
    \small
    \setlength{\tabcolsep}{4pt}
    \begin{tabular}{|l|c|c|c||l|l|l|l|l|}
    \hline
     \multirow{2}{*}{ Data }&\multirow{2}{*}{\begin{tabular}{@{}c@{}}one\\cluster\end{tabular}}&\multicolumn{2}{|c|}{Ours - Gaussian kernel}&\multirow{2}{*}{ GIc }&\multirow{2}{*}{ EM }&\multirow{2}{*}{ kmeans }&\multirow{2}{*}{ Spectral }&\multirow{2}{*}{ Birch }\\
       \cline{3-4} & & \begin{tabular}{@{}c@{}}$0.5n$\\(default)\end{tabular} & \begin{tabular}{@{}c@{}}$0.7n$\\(sensitivity)\end{tabular} &  &  &  &  &   \\ \hline\hline
        Ecoli & 0.2701 & \textbf{0.8762} (4) & 0.8675 (4) & 0.8076 & 0.8467 & 0.8151 & 0.5366 & 0.8308 \\
        Glass & 0.2598 & 0.6267 (6) & 0.5375 (2) & \textbf{0.7123} & 0.6761 & 0.6793 & 0.6681 & -- \\
        Ionosphere & 0.5385 & \textit{0.5363} (6) & \textit{0.5277} (6) & 0.6067 & \textbf{0.7010} & 0.5889 & 0.5401 & 0.5938 \\
        Sonar & 0.4999 & \textbf{0.5048} (8) & 0.5186 (5) & \textit{0.4978} & 0.5006 & 0.5032 & \textit{0.4976} & \textit{0.4976} \\
        Statlog & 0.1425 & 0.8747 (41) & 0.8722 (31) & \textbf{0.9066} & 0.8450 & 0.8116 & 0.1512 & 0.6499 \\
        WDBC & 0.5316 & 0.6893 (5) & 0.7290 (3) & 0.6526 & \textbf{0.8541} & 0.7504 & 0.5423 & 0.5521 \\
        Wine & 0.3380 & 0.7089 (5) & 0.6646 (4) & 0.7128 & \textbf{0.9220} & 0.7187 & 0.6705 & 0.6515 \\
        Yeast & 0.2227 & \textbf{0.7600} (25) & 0.6757 (7) & 0.7385 & 0.6721 & 0.7421 & 0.2711 & 0.7121 \\ \hline
    \end{tabular}
    \caption{Results on the UCI benchmark dataset (Rand Index). The reported configuration of our algorithm is the default maximum boundary $0.5n$; the $0.7n$ column is a sensitivity check on that parameter and not a second configuration. The number in parentheses is the number of clusters our algorithm produced, which is not supplied to it; the reference number of classes is in Table \ref{tab:benchData}, and every other algorithm in this table was given it. The \textit{one cluster} column is the Rand index obtained by placing every point in a single cluster, which requires no algorithm; it is a reference floor and is excluded from the row comparison. Values in italics are below that floor. Bold marks the best value of each row among the default column and the baselines, the sensitivity column being excluded for the same reason. The Birch result for \textit{Glass} is absent from the published benchmark results and is left blank. Each of our columns comes from a single deterministic run of the pipeline described in this article; repeating it reproduces every value exactly.}
    \label{tab:benchUCIresult}
\end{table}
GIc is a recent hierarchical clustering algorithm based on the Genie algorithm (\cite{gagolewski_genie_2016}) and implemented in the genieclust package (\cite{gagolewski_genieclust_2021}). All other algorithms are well known and available on scikit-learn (\cite{pedregosa_scikit-learn_2011}). All algorithms except ours need at least the number of clusters as a parameter. Spectral was used with default parameter $gamma=1$, Birch had default parameters $treshold=0.5$ and $branching\ factor=50$.

The two-class rows should be read with caution. The Rand index counts agreeing pairs and takes a large value under an uninformative partition. The \textit{one cluster} column of Table \ref{tab:benchUCIresult} gives the size of that effect: placing every point in a single cluster scores $0.5385$ on \textit{ionosphere}, $0.5316$ on \textit{WDBC} and $0.4999$ on \textit{sonar}, against $0.14$ to $0.34$ on the other five datasets. On \textit{ionosphere} our algorithm scores below that floor at both boundaries. On \textit{sonar} our default result exceeds it by $0.005$, while GIc, Spectral and Birch fall below it. Those two rows carry no evidence that any method recovered the reference partition, and the differences between methods in them should not be read as standings. A chance-corrected index would be a better instrument on such data.

\label{sec:maxBoundaryEffect}The maximum boundary is the one parameter of the cluster cardinality estimation that is neither scale invariant nor local: it is a rank expressed as a fraction of $n$, so it depends on the size of the whole dataset rather than on the neighborhood of a point. Moving it from $0.5n$ to $0.7n$ changes the Rand index by $+0.040$ on \textit{WDBC} and by $-0.089$ on \textit{Glass}; the eight absolute changes have a median of $0.027$ and five of the eight exceed $0.01$. The mean over the eight datasets is $0.697$ at $0.5n$ against $0.674$ at $0.7n$, so the default is the better of the two settings on average.

The direction of the change is systematic, and it follows from what the parameter asserts. Setting the maximum boundary to $0.5n$ asserts that no cluster contains more than half the data: for a point in a larger cluster the minimum of $\gamma$ is pushed against the edge of the search window, and the rejection test of Algorithm \ref{alg:clusterSizeEstimation} then labels that point's estimate bad and removes it from the mean shift. The last column of Table \ref{tab:benchData} shows that the assertion fails on exactly three of these eight datasets, \textit{ionosphere}, \textit{sonar} and \textit{WDBC}, whose largest class covers $0.641$, $0.534$ and $0.627$ of the sample. On the five datasets where the assertion holds, raising the boundary hurts on all five. On the three where it fails, raising it helps on \textit{sonar} and \textit{WDBC} and does not on \textit{ionosphere}, the one row that stays below the one-cluster floor at either setting. The sign of the change is predicted by the class balance on seven of the eight datasets.

The three datasets where the assertion fails are also the three with the smallest margins over the one-cluster floor in Table \ref{tab:benchUCIresult}: an estimator that cannot represent the majority class cannot recover a two-class partition dominated by it. The default $0.5n$ is also a prior on cluster size. It is far weaker than the number of clusters supplied to the other algorithms in Table \ref{tab:benchUCIresult}, and it is not tuned per dataset, but it remains an assumption about the partition rather than a property of the data. We report the default and do not select between the two columns, since choosing per dataset would require knowledge of the partition we are trying to recover. Determining the maximum boundary algorithmically would remove the assumption along with the parameter.

The number of clusters our algorithm produced is reported in parentheses in Table \ref{tab:benchUCIresult}. Against the reference class counts of Table \ref{tab:benchData}, the algorithm recovers the number exactly on \textit{glass}, produces fewer than the reference on \textit{ecoli}, and produces more on the remaining six, by a wide margin on \textit{statlog} (41 against 7) and \textit{yeast} (25 against 10). The Rand index is tolerant of over-segmentation: splitting a reference class into several clusters leaves every pair from different classes correctly separated and only loses the pairs inside the split class, which is why \textit{statlog} scores $0.875$ with 41 clusters against GIc's $0.907$ with the 7 it was given. On \textit{ionosphere}, \textit{sonar} and \textit{WDBC} the algorithm returns 6, 8 and 5 clusters for two reference classes, which is another view of the failure described above: when the majority class cannot be represented within the maximum boundary, it is recovered as several pieces.

The \textit{yeast} data appears in both comparisons, but the two results are not comparable. Table \ref{tab:tableRen} scores the three largest classes of that dataset under a different preprocessing, and a Rand index computed against three classes has a different floor from one computed against ten.

In general, our algorithm seems competitive with other popular algorithms on the six datasets where the Rand index separates methods, particularly since it can be used without specifying the number of clusters.

\section{Future Work and Conclusion}
The current algorithm represents a first functional prototype and offers numerous opportunities for improvement.
Cardinality estimation is a critical component of our approach, and additional research is required to refine several aspects, including its mathematical formulation, methodological design, and the behavior of the estimator on complex datasets. KDEs could be used after an initial bandwidth ($\sigma$) estimate has been obtained to improve the density estimation. Cluster boundaries could be determined with mode-seeking algorithms rather than via fixed parameters, for example, using one-dimensional mean shift. Furthermore, different parameters and strategies for the mean shift algorithm, such as the gradual increase in area, should also be tested systematically.

In this article, we introduced a new cluster cardinality estimator based on distance distributions and showed how that information can be used by an adaptive mean shift algorithm to cluster real-world datasets without being given the number of clusters. The estimator is scale invariant and local, as shown in Section \ref{sec:gammaDetails}: it carries no constant expressed in units of length, and the parameters applied at a point are determined by its own neighborhood rather than by the extent or the sampling of the rest of the dataset. Three parts of the algorithm fall outside those properties. The maximum boundary is a rank expressed as a fraction of the dataset size; the decision to discard a point's estimate is taken against that boundary and therefore depends on $n$; and the mean shift loop uses a merge threshold and an absolute convergence tolerance rather than the local scale. Removing the maximum boundary would address the first two and is the improvement we would make first.

The proposed cardinality estimator remains a work in progress, with potential for significant improvement and broader applicability beyond clustering tasks. For instance, the cardinality estimator could be applied to a variety of machine learning tasks that require information about small, localized subsets within complex datasets. Currently, some version of a Kernel Density Estimation is often used in such cases. Our method has the advantage of estimating statistics based on distance distributions restricted mainly to a single cluster, and of doing so without a bandwidth. In contrast, KDE methods measure local density in the data space, at a resolution set by a bandwidth that has to be chosen, and often provide limited insight into the overall cluster.

Despite the cardinality estimator being in its early stages, our cluster cardinality-based mean shift algorithm demonstrated competitive clustering performance across a variety of datasets. Within the adaptive mean shift family it is competitive: on the nine datasets of the weighted adaptive mean shift study of \cite{ren_weighted_2014} it obtains the higher Rand index on seven, four of them by a clear margin and three by less than $0.012$, using no knowledge of the number of clusters. That comparison covers one member of the family, and a broader comparison against more recent adaptive mean shift methods is the natural next test of the claim. Against the general clustering benchmark it performed on par with a range of established algorithms, all of which require prior knowledge of the number of clusters. Further improving the cardinality estimator is likely to enhance clustering performance even more.
 
\newpage

\section{Appendix}
\subsection{Algorithms}
\begin{algorithm} [H]
\small
\caption{Cluster Size Estimation}\label{alg:clusterSizeEstimation}
 \hspace*{\algorithmicindent} \textbf{Input} $\mathcal{X},(minBoundary, maxBoundary)$\\
 \hspace*{\algorithmicindent} \textbf{Output} $\mathcal{P},(\mathbf{\hat n},\mathbf{h},\boldsymbol \omega)$
\begin{algorithmic}
\State $j\gets 0$
\For{$\mathbf{x}_i \in \mathcal{X}$}
\State $[y_{(1)},\dots,y_{(n-1)}]\gets $ sort$(||\mathbf{x}_i-\mathcal{X}||)$
\For{$k:minBoundary \rightarrow 1.1maxBoundary$}
    \State $\bar y_{(k)}\gets \sum_1^k y_{(i)}/k$
    \State $s^2_{(k)}\gets \sum_1^k (y_{(i)}-\bar y)^2/k$
    \State $\gamma_{(k)}\gets\frac{s^2}{(\bar y-y_{(k)})^2}$
\EndFor
\State $\hat n_{i1}\gets $argmin$([\gamma_{minBoundary},\dots,\gamma_{maxBoundary}])+minBoundary$
\State $\hat n_{i2}\gets $argmin$([\gamma_{minBoundary},\dots,\gamma_{maxBoundary},\dots,\gamma_{1.1maxBoundary}])+minBoundary$
\If{$\hat n_{i1}=\hat n_{i2}$} \Comment{True if the cluster size estimate is not affected by the maximum boundary}
    \State $j\gets j+1$
    \State $\hat n_j\gets\hat n_{i1}$
    \State $\bar y\gets \sum_{k=1}^{\hat n_j} y_{(k)}/\hat n_j$
    \State $h_j^2 \gets \sum_{k=1}^{\hat n_j} (y_{(k)}-\bar y)^2/\hat n_j$
    \State $\omega_j\gets y_{(\hat n_j)}$
    \State $\mathbf{p}_j\gets\mathbf{x}_i$
\EndIf
\EndFor
\State $\mathcal{P} \gets\{\mathbf{p}_1,\dots\} $ \Comment{Points with good estimate}
\State // parameters of points with good estimate
\State $\mathbf{\hat n}\gets[\hat n_1,\dots]$
\State $\mathbf{h}\gets[h_1,\dots]$
\State $\boldsymbol \omega\gets[\omega_1,\dots]$
\end{algorithmic}
\end{algorithm}

\begin{algorithm}[H]
\small
\caption{Clustering}\label{alg:clustering}
 \hspace*{\algorithmicindent} \textbf{Input} $\mathcal{P}$\\
 \hspace*{\algorithmicindent} \textbf{Output} $\mathcal{M},\mathbf{label}$
\begin{algorithmic}
\State $\mathcal{M}\gets\emptyset$ \Comment{//Mode}
\State $\mathbf{label}\gets\emptyset$ \Comment{//Class of each points}
\For{$\mathbf{p}_i \in \mathcal{P}$}
\State $exist\gets False$
\For{$j:1\rightarrow |\mathcal{M}|$}
\If{$\mathbf{p}_i\approx\mathbf{m}_j$}
\State $exist\gets True$
\State $label_i\gets j$
\State break
\EndIf
\EndFor
\If{$exist==False$}
\State $\mathcal{M}\gets\mathcal{M}\cup\mathbf{p}_i$
\State $\mathbf{label}\gets \mathbf{label}\cup (|\mathcal{M}|)$
\EndIf
\EndFor
\end{algorithmic}
\end{algorithm}

\subsection{Distances in high dimensions} \label{appendix:dim}
The \textit{curse of dimensionality} is often referred to as a reason that distances should not be used in high-dimensional space. \cite{beyer_when_1999} mentions that "as dimensionality increases, the distance to the nearest neighbor approaches the distance to the farthest neighbor". We will show in this section that, under some general conditions, this is a misinterpretation of their demonstration. As the number of dimensions increases, the distance between the nearest and farthest neighbors of an isotropic distribution tends toward a constant that depends on the number of samples in the distribution and not the number of dimensions. This theorem is based on the fact that i.i.d. distributions have a distance distribution that tends towards a normal distribution. The insights we bring in this section contribute to the design of our Gaussian kernel. 

The following theorem approximates Euclidean distance distributions as a normal distribution as the number of dimensions tends toward infinity. \cite{angiulli_behavior_2018} provided the distribution for the squared Euclidean distance; here we derive the Euclidean distance.
\begin{theorem}\label{theorem:highDimDistanceDist}
Let $\mathbf{Y}_d$ be any multivariate random distribution such that $\mathbf{Y}_d=[Y_1, Y_2,\dots,Y_d], Y_i\in \mathds{R}$, all $Y_i$ are i.i.d. and $\mathrm{E}[Y_i^4]<\infty$. Then the distance distribution $X_d=||\mathbf{Y}_d||$ converges in distribution to a normal distribution such that 
\begin{equation} \label{eq:chiLimit}
    \lim_{d \to \infty} X_d  \xrightarrow{dist}\ N(\sqrt{d\mu_{Y^2} },\tfrac{\sigma_{Y^2}^2}{4\mu_{Y^2}})
\end{equation}
where $\mu_{Y^2}$ is the expectation of $Y_i^2$ and $\sigma_{Y^2}^2$ is the variance of $Y_i^2$.
\end{theorem}
\begin{proof}
We start by proving it for $X_d^2$ and then for $X_d$. $X_d^2$ tends in distribution toward a normal distribution following the central limit theorem (CLT) such that when $d$ tends toward infinity:
\begin{align*}
    \frac{\sqrt{d} \left (\tfrac{\sum Y_i^2}{d}-\mu_{Y^2}\right )}{\sigma_{Y^2}}&\xrightarrow{dist} \mathcal{N}(0,1) \\
   \sum_{i=1}^d Y_i^2-\mu_{Y^2} d &\xrightarrow{dist}  \mathcal N (0,d\sigma_{Y^2}^2) \\
    \sum_{i=1}^d Y_i^2 &\xrightarrow{dist} \mathcal{N}(d\mu_{Y^2} ,d\sigma_{Y^2}^2)
\end{align*}
For example, when $Y_i\sim \mathcal{N}(0,1)$, $X_d^2$ is a chi-squared distribution with $Y_i^2$ having $\mu_{Y^2}=1$ and $\sigma_{Y^2}^2=2$, which makes $X_d^2\sim \mathcal N(d,2d)$ when d is large.

The proof for the square of the distance makes use of the delta method, which states that if there is a sequence of random variables $X_d$ satisfying

\begin{equation*}
    {\sqrt{d}[X_d-\theta]\,\xrightarrow{dist}\,\mathcal{N}(0,\beta^2)}
\end{equation*} 
, then
\begin{equation*}
    \sqrt{d}[g(X_d)-g(\theta)]\,\xrightarrow{dist}\,\mathcal{N}(0,\beta^2[g'(\theta)]^2)
\end{equation*}

for any function $g$ satisfying the property that $g'(\theta)$ exists and is non-zero valued. Continuing from the CLT demonstration:
\begin{align*}
    \sqrt{d}\left (\tfrac{\sum Y_i^2}{\sqrt{d}}-\mu_{Y^2}\sqrt{d}\right )\,&\xrightarrow{dist}\,\mathcal{N}(0,\sigma_{Y^2}^2d)\\
    \intertext{let $g(x)=\sqrt{x}$, $\theta=\mu_{Y^2}\sqrt{d}$ and $\beta^2=\sigma_{Y^2}^2d$ with $[g'(\theta)]^2=\tfrac{1}{4\mu_{Y^2} \sqrt{d}}$, then:}
    \sqrt{d}\left (\tfrac{\sqrt{\sum Y_i^2}}{d^{1/4}}-\sqrt\mu_{Y^2} d^{1/4}\right )\,&\xrightarrow{dist}\,\mathcal{N}(0,\tfrac{\sigma_{Y^2}^2\sqrt{d}}{4\mu_{Y^2}})\\
    \sqrt{\sum_{i=1}^d Y_i^2}-\sqrt{\mu_{Y^2} d} \,&\xrightarrow{dist}\,\mathcal{N}(0,\tfrac{\sigma_{Y^2}^2\sqrt{d}}{4\mu_{Y^2}})\tfrac{1}{d^{1/4}}\\
        \sqrt{\sum_{i=1}^d Y_i^2} \,&\xrightarrow{dist}\,\mathcal{N}(\sqrt{d\mu_{Y^2} },\tfrac{\sigma_{Y^2}^2}{4\mu_{Y^2}})\\
\end{align*}
Which proves the theorem.
\end{proof}

Theorem \ref{theorem:rangeHighDim} shows that distances between neighbors of an i.i.d. distribution tend toward a constant in high dimensions:
\begin{theorem}\label{theorem:rangeHighDim}
Let $\mathbf{Y}_d= [Y_1,\dots,Y_d ]$ be an i.i.d. random vector with $\mathrm{E}[Y_i^4]<\infty$, and let $X_{(1:n)},\dots,X_{(n:n)}$ be the order statistics of $n$ independent copies of $X=||\mathbf{Y}_d||$. Then
\begin{equation}
    \lim_{d\rightarrow\infty}\mathrm{E}[X_{(n:n)}-X_{(1:n)}]=c
\end{equation}
, where $c$ is a constant independent of $d$ for a given $Y_i$ and $n$.
\end{theorem}
\begin{proof}
    From theorem \ref{theorem:highDimDistanceDist}, $\lim_{d \to \infty} X  \xrightarrow{dist}\ N(\sqrt{\mu d},\tfrac{\sigma^2}{4\mu})$, writing $\mu=\mu_{Y^2}$ and $\sigma^2=\sigma_{Y^2}^2$. Let $A\sim N(\sqrt{\mu d},\tfrac{\sigma^2}{4\mu})$ and $B\sim N(0,\tfrac{\sigma^2}{4\mu})$. Then $A=B+\sqrt{\mu d}\ $ and $A_{(i:n)}=B_{(i:n)}+\sqrt{\mu d}$. Now we calculate the expectation of the difference between the farthest and the closest neighbor
    \begin{align*}
        \mathrm{E}[A_{(n:n)}-A_{(1:n)}]&={E}[(B_{(n:n)}+\sqrt{\mu d})-(B_{(1:n)}+\sqrt{\mu d})]\\
        &={E}[B_{(n:n)}-B_{(1:n)}]
    \end{align*}
    The quantity ${E}[B_{(n:n)}-B_{(1:n)}]$ is the expectation of the range (\cite{david_order_2003}) and can be calculated numerically easily for any distribution and $n$ value. Since it is independent of $d$, the expectation of the range of the approximation is only dependent on $\sigma^2$, $\mu$, and $n$ and not the number of dimensions.
\end{proof}

Because the distance distribution converges in distribution to a normal distribution with a variance that is independent of the dimension, the difference between the farthest and closest neighbor also converges, even if the mean of the distribution increases.

This brings context to \cite{beyer_when_1999} where they looked at the ratio of distances between the closest and farthest neighbor. If $\sqrt{\mu d}+B_{(1)}$ represents the closest neighbor, then the ratio of distances when $d$ tends toward infinity can be written in terms of order statistics:
\begin{equation}
    \lim_{d \to \infty} \frac{\text{E}[\sqrt{\mu d}+B_{(1)}]}{\text{E}[\sqrt{\mu d}+B_{(n)}]}= 1
\end{equation}
since
\begin{equation*}
    \lim_{d \to \infty} \sqrt{\mu d}=\infty\quad \text{and} \quad \mathrm{E}[B_{(1)}],\mathrm{E}[B_{(n)}]\in\mathbb{R}
\end{equation*}

This confirms the findings of \cite{beyer_when_1999} for i.i.d. distributions, but also adds context. They mentioned that "as dimensionality increases, the distance to the nearest neighbor approaches the distance to the farthest neighbor". The ratio of distances tends toward 1, but the expected distance between closest and farthest neighbor can stay the same, as is the case for any multivariate distribution with i.i.d. distributions along all axes. This also confirms findings in \cite{aggarwal_surprising_2001} where they found that the distance between farthest and nearest neighbor tends toward a constant for the Euclidean norm.

\subsection{Euclidean Distances} \label{sec:euclidean_distance}
The distribution of Euclidean distances from the mean in normal isotropic multivariate distributions follows a chi distribution with $d$ degrees of freedom and scale parameter $\sigma$. This version of the chi distribution is uncommon, but is referenced in \cite{johnson_continuous_1994} (p. 452); it is also the scale family of the chi distribution (\cite{casella_statistical_2001} (p. 116)). The chi distribution is related to the normal distribution by the following Theorem:

\begin{theorem}\label{theorem_distance_chi}
Let $\mathbf{Y}\in\mathds{R}^d$ be any random variable following an isotropic multivariate normal distribution such that its probability density function (pdf) is:
\begin{equation}\label{eq:pdf_iso_theorem}
    f_\mathbf{Y}(\mathbf{y})=\frac{1}{\sqrt{(2\pi \sigma^2)^d}}e^{ \left (-\frac{||\textbf{y}||^2}{2\sigma^2} \right )}.
\end{equation}
The distribution of Euclidean distances $X=||\mathbf{Y}||$ is the chi distribution with $d$ degrees of freedom and scale parameter $\sigma$ with pdf
\begin{align*}  
    f_X(x) &=  \frac{1}{\Gamma(\frac{d}{2})2^{(d/2)-1}\sigma^d}
       e^{-x^2/2\sigma^2}x^{d-1}\\
       \intertext{and cumulative density function (cdf)}
       P(X<x)&=\frac{\gamma(\frac{d}{2},\frac{x^2}{2\sigma^2})}{\Gamma(\frac{d}{2})}.
\end{align*}
\end{theorem}

\begin{proof} 
We will start with the multivariate normal distribution (eq.\ref{eq:PDF_multivariate}) and modify it to obtain the multivariate isotropic Gaussian distribution (eq.\ref{eq:pdf_iso_theorem}). The multivariate normal distribution is
\begin{equation} \label{eq:PDF_multivariate}
f(\textbf{x}:\boldsymbol\mu,\Sigma)= \frac{1}{\sqrt{|\Sigma|(2\pi)^d}}\exp{(-\frac{1}{2}(\mathbf{x}-\boldsymbol\mu)^{\textit{T}}\Sigma^{-1}(\mathbf{x}-\boldsymbol\mu))},
\end{equation}

where $ \boldsymbol\mu \in \mathds{R}^d$ is the mean, $\Sigma$ is a square covariance matrix, $d$ is the number of dimensions and $|\Sigma|$ is the determinant of the covariance matrix. \par
Using an average of $\boldsymbol\mu=\boldsymbol0$ removes the term without loss of generality since we want to calculate our cdf from the center of the distribution. To get an isotropic normal density, we reduce the covariance matrix to a diagonal matrix with value $\sigma^2$ on the diagonal. This leads to
\[\textbf{x}^{\textit{T}}\Sigma^{-1}\textbf{x}=\frac{||\textbf{x}||^2}{\sigma^2}\]
and 
\[|\Sigma|=(\sigma^2)^d.\]
 We then obtain:
\begin{equation} \label{eq:PDF_iso}
    f(\textbf{x}:\sigma)= \frac{1}{(\sqrt{2\pi}\sigma)^d}e^{-||\textbf{x}||^2/2\sigma^2},
\end{equation}
which is eq.\ref{eq:pdf_iso_theorem} from theorem \ref{theorem_distance_chi}.

In eq.\ref{eq:PDF_iso}, $\mathbf{x}$ is a vector, but since we only use it's norm, it can be used as a variable instead of $\mathbf{x}$ without losing generality. This results in an univariate function of the radial distance $r$ instead of a multivariate one:
\begin{equation} \label{eq:PDF_Iso_univariate2}
    f(r:\sigma)= \frac{1}{(\sqrt{2\pi}\sigma)^d}e^{-r^2/2\sigma^2},
\end{equation}
where $||\textbf{x}||^2=r^2$. 
\par
We now want to find the cdf of the distribution $X$ of distances from the center of an isotropic multivariate Gaussian distribution. It follows this form $P(X<r;d,\sigma)$: the probability of choosing a point at random from a normal distribution with a distance from the center of $r$ or less. We count the number of points at every distance $r$ and calculate their combined probability density function. To calculate this, we construct an hypersphere with a Gaussian density and calculate its mass. The d-dimensional multivariate Gaussian density defines iso-contours of equiprobable points. In the case of an isotropic multivariate Gaussian density, these lie on the surface of a hypersphere in $\mathds{R}^d$.
\par

The mass of the d-dimensional hypersphere can be calculated with the following formula:
\[M(r)=\int_{0}^{r} density(r)\ dV\]
For the hypersphere density we use eq.\ref{eq:PDF_Iso_univariate2}.
Which results in the cdf:
\begin{equation}\label{eq:cdf_r_Iso}
F(r)=P(X<r)=\int_{0}^{r}\frac{1}{(\sqrt{2\pi}\sigma)^d}e^{-\frac{x^2}{2\sigma^2}} dV,
\end{equation}
where $V$ is the $d$-dimensional volume (content) of the hypersphere. Let $S_d$ be the hyper-surface area of a d-dimensional hypersphere of unit radius. From the wolfram\footnote{\url{https://mathworld.wolfram.com/Hypersphere.html}}:
\begin{equation*} 
    S_d= \frac{2\pi^{d/2}}{\Gamma(\frac{d}{2})}
  \end{equation*}
  and
 \[V=\int_{0}^{r}S_d x^{d-1}dx=\frac{S_dr^d}{d}. \]
Which we use to calculate $dV$:
\begin{equation}\label{eq:dV}
\begin{split}
        V &= \frac{2\pi^{d/2}r^d}{d\Gamma(\frac{d}{2})} \\
        dV &=\frac{dV}{dr}dr= \frac{2\pi^{d/2}r^{d-1}}{\Gamma(\frac{d}{2})}dr
\end{split}
\end{equation}
\par
We can now combine eq.\ref{eq:cdf_r_Iso} and \ref{eq:dV} to calculate the cdf:
\begin{align}
         P(X<r) &= \int_0^r \frac{2\pi^{d/2}}{\Gamma(\frac{d}{2})}
       \frac{1}{(\sqrt{2\pi}\sigma)^d}e^{-x^2/2\sigma^2}x^{d-1}\ dx \label{eq:cdf_r_init}, \\
       \intertext{we move the terms we can out of the integral, }
       &= \frac{2}{2^{d/2}\Gamma(\frac{d}{2})\sigma^d}
        \int_0^r e^{-x^2/2\sigma^2}x^{d-1}\ dx \nonumber, \\
        \intertext{then modify the terms of the integral in order to obtain the lower gamma function $\gamma$ starting with a variable substitution:}
        &\frac{x^2}{2\sigma^2}=y \xrightarrow[]{}
        \left \{
  \begin{aligned}
    &\frac{x}{\sigma^2}dx=dy \\
    &x=\sqrt{2y\sigma^2}.\\
  \end{aligned} \right. \nonumber
        \intertext{The upper limit of the integral goes from $r$ to $\frac{r^2}{2\sigma^2}$:} 
        &=\frac{2\sigma^2}{2^{d/2}\Gamma(\frac{d}{2})\sigma^d}\int_0^r e^{-x^2/2\sigma^2}x^{d-2} \frac{x}{\sigma^2}\ dx = \frac{2\sigma^2}{2^{d/2}\Gamma(\frac{d}{2})\sigma^d} \int_0^\frac{r^2}{2\sigma^2} e^{-y} (\sqrt{y2\sigma^2})^{d-2}\ dy \nonumber\\
        &=\frac{2^{d/2}\sigma^d}{2^{d/2}\Gamma(\frac{d}{2})\sigma^d}\int_0^\frac{r^2}{2\sigma^2} e^{-y} y^{d/2-1} \ dy\ \  \nonumber\\
        \intertext{By definition of the $\gamma$ function we obtain:}
        &=\ \frac{1}{\Gamma(\frac{d}{2})}\ \gamma(\frac{d}{2},\frac{r^2}{2\sigma^2})\nonumber\\
        P(X<r)&=\frac{\gamma(\frac{d}{2},\frac{r^2}{2\sigma^2})}{\Gamma(\frac{d}{2})} \label{eq:Chi_cdf_sigma}
\end{align}

You can change variable $r$ to $x\sigma$, where $x$ is the Mahalanobis distance:\\
\begin{equation} \label{cdfManalanobis}
 P(X<x)=\frac{\gamma(\frac{d}{2},\frac{x^2}{2})}{\Gamma(d/2)}
\end{equation}

Eq.\ref{eq:Chi_cdf_sigma} is the cdf of the Chi distribution. It is related to the chi-square distribution, which is the distribution of the sum of $d$ squared standard normal distribution. We will work with eq.\ref{eq:Chi_cdf_sigma} instead to keep generality. For the rest of this document we call the distribution having this cdf the Chi scale family distribution and denote it this way: $X\sim\chi(d,\sigma)$.

We can obtain the pdf of the chi distribution from eq.\ref{eq:cdf_r_init} by deriving it:

\begin{equation} \label{eq:PDF_r_iso} 
    f(x) =  
       \frac{1}{2^{d/2-1}\Gamma(\frac{d}{2})\sigma^d}e^{-x^2/2\sigma^2}x^{d-1}\ 
\end{equation}

We now concluded that the distribution of distances to the center of an isotropic Gaussian distribution follows a chi scale family distribution.
\end{proof}

\begin{theorem}\label{theo:point-to-point_euclidean}
Let $\mathbf{Y},\mathbf{Q}\sim\mathcal{N}(\boldsymbol\mu,\sigma^2I_d)$ be independent and $X=||\mathbf{Y}-\mathbf{Q}||$. Then, $X$ is a distribution of Euclidean distances between 2 points randomly sampled from $\mathcal{N}(\boldsymbol\mu,\sigma^2I_d)$ and follows a chi distribution with d degrees of freedom  and scale parameter $\sigma\sqrt{2}$ (variance $2\sigma^2$) noted $\chi(d,\sigma\sqrt{2})$.
\end{theorem}
\begin{proof}
By the sum of independent normal random variable property, if $\mathbf{A}=\mathbf{Y}-\mathbf{Q}$, then
$\mathbf{A} \sim \mathcal{N}(0,2\sigma^2I_d)$, whose components have standard deviation $\sigma\sqrt{2}$, and by theorem \ref{theorem_distance_chi} $||A||\sim\chi(d,\sigma\sqrt{2})$.
\end{proof}

Theorem \ref{theo:point-to-point_euclidean} confirms results presented in \cite{thirey_distribution_2015} where they found that the distance distribution between points of a normal distribution with parameter $\sigma=1$ follows a distribution with pdf:
\begin{equation}
    f(x,d)=\frac{2^{1-d}}{\Gamma\left(\frac{d}{2}\right)}\exp{\left(\frac{-x^2}{4}\right)}x^{d-1}
\end{equation}
This is the same result as using Theorem \ref{theorem_distance_chi} and theorem \ref{theo:point-to-point_euclidean} to calculate the point-to-points distance pdf in a Gaussian distribution with $\sigma=1$:
\begin{equation}
    f(x,d)=\frac{1}{\Gamma\left(\frac{d}{2}\right)2^{(d/2)-1}\sqrt{2}^d}
    \exp{\left(\frac{-x^2}{2\sqrt{2}^2}\right)}x^{d-1}
\end{equation}
which is the pdf of the chi($d,\sqrt 2$) distribution. Apart from confirming previous results, theorem \ref{theo:point-to-point_euclidean} also links the interpoint distance distribution with the chi distribution and defines the relation between the sigma parameter of the normal distribution and its effect on the interpoint distance distribution.

\printbibliography

\end{document}